\documentclass[letterpaper, 10 pt, journal, twoside]{IEEEtran}

\bstctlcite{bstctl:etal, bstctl:nodash, bstctl:simpurl}

\usepackage{times}
\usepackage{graphicx}
\usepackage{amsmath}
\usepackage{amssymb}
\usepackage{textcomp}
\usepackage{multirow}
\usepackage{booktabs}
\usepackage{flushend}
\usepackage{xcolor}
\usepackage{multicol}
\usepackage{color}
\usepackage{threeparttable}
\usepackage{makecell}
\usepackage{bbding}
\usepackage{array}
\usepackage{pifont}
\usepackage{colortbl}

\usepackage{epsfig}

\usepackage{textcomp}
\usepackage{adjustbox}

\usepackage{hhline}
\usepackage{subcaption}
\usepackage{enumitem}
\usepackage{verbatim}
\usepackage{multirow}
\usepackage{graphicx}
\usepackage{%geometry,
    tabularx, % for 'tabularx' env.
    ragged2e, % for '\RaggedRight' macro
    booktabs, % '\toprule' '\midrule', etc. 
    amssymb}  % for '\checkmark' macro
\newcolumntype{L}{>{\RaggedRight\arraybackslash\hspace{0pt}}X}
\usepackage[noadjust]{cite}

\usepackage{gensymb}

\usepackage{wrapfig}

\newcommand{\ins}{\textcolor{black}} %mark changes in script in text color
\usepackage[figureposition=bottom,tableposition=top]{caption}
\def\x{{\mathbf x}}

\usepackage[colorlinks,pagebackref=false,citecolor=blue,bookmarks=false,hypertexnames=true]{hyperref}

\begin{document}

\bstctlcite{IEEEexample:BSTcontrol}

\title{
Surround-View Fisheye Optics in Computer \\ Vision and Simulation: Survey and Challenges
}

\author{
Daniel Jakab,  
Brian Michael Deegan,
Sushil Sharma,
Eoin Martino Grua,
Jonathan Horgan,\\
Enda Ward,
Pepijn Van De Ven,
Anthony Scanlan, % <-this % stops a space 
and
Ciarán Eising~\IEEEmembership{Senior Member,~IEEE} \\

%\thanks{Manuscript received XX XXX 2023; revised XX XXX 2023; accepted XX XXX 202X. Date of publication XX XXX 202X; date of current version XX XXX 202X. The Associate Editor for this article was XXXXXXXX. (Corresponding author: Daniel Jakab (daniel.jakab@ul.ie).}
\thanks{D. Jakab, S. Sharma, E. M. Grua, P. Van De Ven, A. Scanlan, and C. Eising are with the Department of Electronic and Computer Engineering, University of Limerick, Limerick, V94 T9PX, Ireland.}
\thanks{B. Deegan is with the Department of Electrical and Electronic Engineering, University of Galway, Galway, H91 HX31, Ireland.}
\thanks{D. Jakab, P. Van De Ven, A. Scanlan, C. Eising, and B. Deegan are also with Lero (the Science Foundation Ireland Research Centre for Software), University of Limerick, Limerick, V94 T9PX, Ireland.}
\thanks{J. Horgan and E. Ward are with Valeo Vision Systems, Dunmore Rd., Tuam, Co. Galway, H54 Y276, Ireland.}
\thanks{This work is co-funded by Lero the Science Foundation Ireland Research Centre for Software (Grant Number 13/RC/2094\_P2) and Valeo Vision Systems.}
}

%\markboth{IEEE Transactions on Intelligent Transportation Systems}
%{Jakab \MakeLowercase{\textit{et al.}}: Surround-View Fisheye Optics in Computer Vision and Simulation: Survey and Challenges}

\maketitle

\begin{abstract}
In this paper, we provide a survey on automotive surround-view fisheye optics, with an emphasis on the impact of optical artifacts on computer vision tasks in autonomous driving and ADAS. The automotive industry has advanced in applying state-of-the-art computer vision to enhance road safety and provide automated driving functionality. When using camera systems on vehicles, there is a particular need for a wide field of view to capture the entire vehicle's surroundings, in areas such as low-speed maneuvering, automated parking, and cocoon sensing. However, one crucial challenge in surround-view cameras is the strong optical aberrations of the fisheye camera, which is an area that has received little attention in the literature. Additionally, a comprehensive dataset is needed for testing safety-critical scenarios in vehicle automation. The industry has turned to simulation as a cost-effective strategy for creating synthetic datasets with surround-view camera imagery. We examine different simulation methods (such as model-driven and data-driven simulations) and discuss the simulators' ability (or lack thereof) to model real-world optical performance. Overall, this paper highlights the optical aberrations in automotive fisheye datasets, and the limitations of optical reality in simulated fisheye datasets, with a focus on computer vision in surround-view optical systems.
\end{abstract}

\IEEEpeerreviewmaketitle

\section{Introduction}
\label{sec:introduction}
%- Computer Vision, a relatively new field
%- Use of simulation in computer vision is new
%- training models in simulation and deploying into real-world (this is our problem)
%- optical artifacts/lack of in simulation 
%- Outline what did I do?
%- concluded by outlining structure of paper in sections
\begin{figure}[t]
    \centering
    \includegraphics[width=\columnwidth]{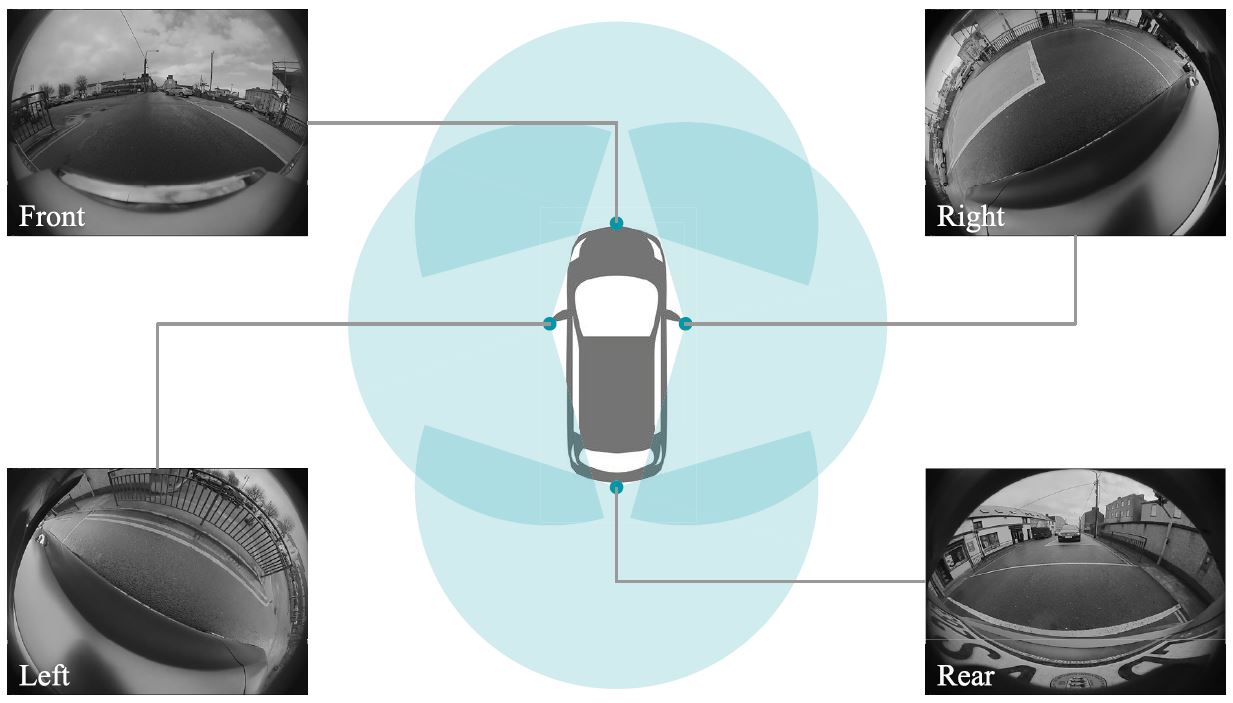}
    \caption{{\bf Illustration of a surround-view camera system for automated driving}. The cameras are FV (top left), MVR (top right), MVL (bottom left), and RV (bottom right) \cite{kumar2023surround}.}
    \label{fig:sv-sys}
\end{figure}

%- Computer Vision, a relatively new field
Computer Vision is a field of research where algorithms are designed to computationally interpret the scene from images of the real world that the camera is in. With the coming to the fore of neural networks in the last decade, computer vision research has grown exponentially. Automated driving is an industry where computer vision has begun to dominate \cite{CGV-079}. In parallel, surround-view camera systems have received attention. Through a combination of several fisheye cameras located on a vehicle, such camera systems offer a full 360$^\circ$ view around the vehicle, as illustrated in Figure~\ref{fig:sv-sys}. Surround-view systems were traditionally employed for scene-viewing applications (such as blind zone monitoring \cite{hughes2009} and bird-eye view vision \cite{birdeye2008}). More recently, there has been particular interest in the computer vision tasks that can be undertaken by surround-view systems, developing from low-speed applications several years ago \cite{HEIMBERGER201788} to more complete perception tasks in vehicle autonomy \cite{deng2019restricted, kumar2023surround}.
\begin{figure*}[t]
    \centering
    \includegraphics[width=7in, keepaspectratio]{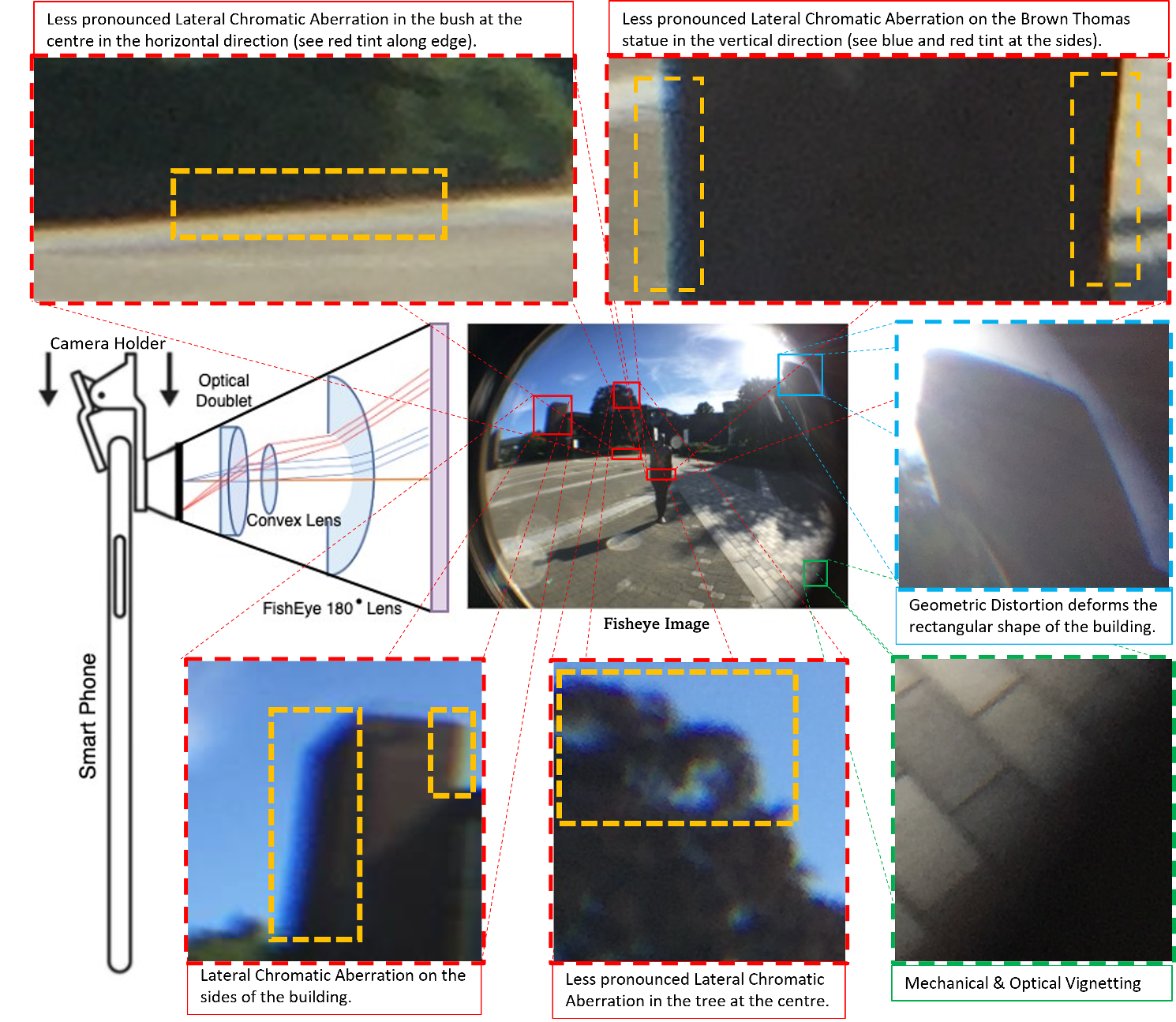}
    \caption{\ins{{\bf Illustration of 180° fisheye lens/camera combination}. The captured image shows many optical artifacts associated with fisheye cameras, which we will discuss later in the paper. Note: the marked dotted boxes (orange) represent lateral chromatic aberration (red) and strong astigmatism or optical blur,  geometric distortion effects on a building (blue), and both mechanical and optical vignetting at the periphery with slight shadow (green).}}
    \label{fig:fisheye}
\end{figure*}
Part of this field of research is the creation of datasets of images from automotive scenes using specific camera and lens combinations. These datasets are the basis for developing, training, and deploying computer vision models in the real world. However, due to variations in optics and electronics (both in design and due to manufacturing tolerances), different cameras capture different scenes in different ways. 

\ins{The impact of the electronic Image Signal Processing (ISP) on the performance of computer vision is discussed in detail in \cite{jimaging5100078, jimaging9120260}. The impact of the variation in optics on computer vision performance is, perhaps, not well studied in a single place. In particular, wide fisheye cameras tend to suffer more from optical artifacts and are much more apparent in visual imagery than in other camera types\cite[p.233]{Hecht2017}. Primarily, the large incidence angle of light due to field curvature in fisheye systems leads to much more serious astigmatism and a higher-order chromatic aberration, which is difficult to correct especially as the FOV angle increases \cite{fan2021calculation, martin2004design}. Most papers only discuss strategies to compensate or correct for optical artifacts which is much more apparent in fisheye or wide FOV panoramic optical systems  \cite{gao2022compact, huang2017design, martin2004design}. In contrast, in this paper, we discuss the optical artifacts associated with wide FOV systems and their impact on computer vision performance.}

\ins{Fisheye cameras are unique and not well understood in the computer vision space as a result of the extreme optical artifacts present in the images \cite{fan2021calculation, pontinen2012study, martin2004design}. For this reason, more experiments are needed to understand how computer vision performance is affected by the optical artifacts of this type of camera lens.} One can consider that optical artifacts increase the further off-axis a light ray is. With the wide optics of fisheye, there is, therefore, a greater potential for optical artifacts. Thus, in this paper, we will discuss what is known about the impact of optical aberrations on computer vision performance in surround-view systems.

For safety-critical applications, such as vehicle autonomy, this lack of understanding of computer vision model behavior with different lenses is potentially dangerous, especially if all potential scenes cannot be investigated. Outcomes on the road can only be investigated by examining different traffic scenarios, including corner cases that rarely occur and are the most challenging to reproduce in test facilities.
Simulation can be a powerful tool where any corner-case traffic scenario can be reconstructed as additional data for training computer vision models. In this paper, we will survey the popular vision simulation methods, and pay particular attention to their ability to simulate surround-view cameras accurately, including fisheye geometry and other optical artifacts. Not only should the optical artifacts in simulations be accurate for training, but if we can adapt the parameters of a simulated optical model, it would be possible to show how computer vision models behave using different camera lenses or cameras produced within varying manufacturing tolerances. However, despite its popularity in automated driving research, the use of simulation in this area is a relatively new development. For the particular use case of surround-view systems, significant further development is required for simulations to approach photorealism.

%- training models in simulation and deploying into real-world (this is our problem)
Since simulation is identified as a potential solution to the scarcity of well-sampled fisheye data, a problem exists where computer vision models should be trained with simulation data and deployed into the real world. However, as will be shown in this paper it is not as simple as applying fisheye distortion to simulation. The entire lens as well as its realistic effects on real-world images should be modelled in simulation. For example, Figure~\ref{fig:fisheye} shows a fisheye image with many readily apparent optical artifacts. \ins{For example, mechanical and optical vignetting is visible at the corners of the image where there is a slight shadow cast on the pixels. The effects of geometric distortion can be seen as the shape of the building is deformed from being rectangular to a convex-like shape. Lateral chromatic aberration is evident on the sides of buildings and trees. In particular, the left outline of the building exhibits a deep blue color, which contrasts with the light blue sky (see Section \ref{sec:ca-sys} Figure \ref{fig:fisheye-lateral-CA}). The blurriness of the buildings and scenery constitutes both sagittal and tangential astigmatism.}

The lack of optical artifacts in simulation is a shortcoming because the effects caused by optical models are a crucial part of the real world. If these are not simulated, there is a lack of information provided to computer vision models in training, leading to potentially unpredictable performance on the road. Therefore, in current circumstances, computer vision models cannot learn the full scope of the real world in simulation and arguably cannot be deployed completely safely as a result.

%- Outline what did I do?
%other reviews
There have been other surveys in the field discussing fisheye cameras for automotive systems and simulation. None of them has a specific focus on the optical background for automotive surround-view systems and the optical artifacts that are created by the optical systems, other than the obvious geometric distortion. An early survey \cite{hughes2009} focused on the scene-viewing application of blind zone monitoring and discussed the effects of fisheye geometric distortion and light fall-off.
In \cite{eising2021near}, a part-survey, part-positional argument is provided on how surround-view perception systems should be structured, based on what the authors coined the 4Rs of automotive surround-view computer vision (Reconstruction, Recognition, Relocalization, and Reorganization), itself based on earlier work known as the 3Rs of computer vision \cite{MALIK20164}.
In Kumar et al. \cite{kumar2023surround}, a review of optical systems in automated driving for fisheye (i.e. fisheye projection models, the surround-view camera system, perception tasks, and popular automotive datasets) is conducted.
In \cite{9913937}, a superficially similar survey to \cite{kumar2023surround} is provided. However, in \cite{9913937}, the authors provide more detail on automotive fisheye datasets (both real and simulated), and the three tasks of object detection and tracking, semantic segmentation, mapping and localization, and surround-view monitoring. Other fisheye-related surveys such as Gao et al. \cite{gao2022review}, focus on the different types of wide field-of-view (FOV) systems for 360$^\circ$ panoramic, catadioptric as well as briefly discussing fisheye segmentation for autonomous driving. Finally, Mutsch et al \cite{mutsch2023model} provide an insightful overview of the advantages and drawbacks of different simulator types for autonomous driving, where an important insight into the different types of simulation available, such as model-driven and data-driven simulation, is provided. A lot of crucial points on simulation are raised in Mutsch's work which will be investigated here from the surround-view camera perspective. 

In all of the above surround-view surveys, only the obvious geometric distortion is surveyed in detail, and as such, we only mention this topic briefly in this paper. In none of the above are the other optical artifacts of surround-view cameras discussed (except for \cite{hughes2009} that discusses light fall off). While simulations are discussed in these surveys as a means to augment datasets for automated driving development, there is also no discussion provided of how realistic the optics of the simulations are.  In this paper, we aim to provide this discussion. Therefore, to complement the other surveys mentioned, this paper will (1) elaborate specifically on the types of image quality effects found in fisheye optical systems, (2) investigate how these optical artifacts impact computer vision, (3) establish the current trend in simulation frameworks, distinguishing between different simulation approaches applicable for automated driving and considering the accuracy of optical artifacts in current simulations. 

%- concluded by outlining structure of paper in sections
This paper is organized as follows. In Section \ref{sec:fish-opt-sys}, we will briefly discuss what surround-view optical systems are and how they are designed.
In Section \ref{sec:optical-effects}, we will discuss the main image quality factors including Chromatic Aberration, Astigmatism, Vignetting, and Geometric Distortion specifically for fisheye cameras and how these affect computer vision. 
Then we examine computer vision applied to all surround-view systems in Section~\ref{sec:compvis}, including surveying some recent new proposals on how to measure computer vision performance considering the spatially variant optics of surround-view cameras.
Finally, in Section \ref{sec:sim}, an in-depth discussion is made where the photorealism and image quality of simulations are discussed and how fisheye can become part of computer simulators (i.e. CARLA, Unity, etc.).

\section{Fisheye Optical Design}
\label{sec:fish-opt-sys}
The optical design of a fisheye lens is a significant undertaking, and we do not intend to cover all the details here. This is the topic of papers (e.g., \cite{fan2019design, samy2015simplified}) and books (e.g., \cite{Hecht2017}) in itself. The aim here is to give a reader some concept of the complexity of the design and to highlight that, while much of the optical design is to combat optical artifacts, it is impossible to completely remove such artifacts.

The goal of an optical camera system is to form a sharp image on the image sensor. We can geometrically consider this as light rays converging on an image plane, as demonstrated in Figure~\ref{fig:fisheye}, where we show a fisheye image captured by a fisheye lens. The optical lens illustrated in Figure~\ref{fig:fisheye} shows a simple three-lens system, though fisheye lenses commonly have more elements. The system shown consists of (1) a fisheye lens, i.e., a lens with a very wide FOV; (2) a convex lens where light rays converge upon exiting; and (3) an optical doublet which is a lens consisting of two different optical media and is typically found in fisheye camera systems. In an optical doublet, the two different media are typically two different types of glass material fused together, which aids in the reduction of chromatic aberration in lenses (discussed in further detail in Section~\ref{sec:ca-sys}). Naturally, fisheye optical systems can become more complex when additional lenses are added to this system depending on its application.
In \cite[259-266]{Hecht2017}, further detail is introduced about ray tracing by modeling optical surfaces as matrices.

Ray tracing is a method for modeling the projection of a virtual ray (representing the light ray) through an optical system, analyzing the refraction of light as it enters different lens media (OpticStudio\footnote{\label{ops}Zemax OpticStudio: \url{https://www.zemax.com/pages/opticstudio}} is an example of such a ray-tracing software tool). An optical system has virtual light rays that should ideally converge at a specific point at the end of an optical system. Both the thickness and the refractive index of the medium affect the direction of virtual rays. In an optical system different types of media can be placed (i.e., lenses with different refractive indices) together using Snell's law \cite[259]{Hecht2017}. Each lens has an individual power associated with it which is the reciprocal of its focal length as described by the equation:
\begin{equation}
    \phi_i = \frac{1}{f_i}
\end{equation}
Where $f_i$ is the $i^\text{th}$ focal length of a lens in an optical system where there can be $n$ optical lenses.
A typical fisheye lens (which is the large disc-shaped lens in Fig. \ref{fig:fisheye}) has very high optical power due to its very small focal length leading to the extreme convergence of light. Abnormally high optical powers lead to extreme optical aberrations as light bends at a wide angle through media. It is difficult to remove the optical artifacts caused by extreme lens powers. Optical Artifacts that directly affect optical systems such as Chromatic Aberration, Astigmatism, and Vignetting are discussed in Section \ref{sec:optical-effects}. The power tends to increase as more convex lenses are added to an optical system. However, not all lenses in an optical system are positive (e.g., the optical doublet shown in Fig. \ref{fig:fisheye}, in which the convex part of the lens is positive (converging) whereas the concave part is negative (diverging)). The overall lens can be positive if the convex part has higher power than the diverging part.
The total power of an optical system (reciprocal of the focal length), can be described as:
\begin{equation} \label{eq:power}
\Phi = \phi_1 + \phi_2 +...+ \phi_i +...+ \phi_n
\end{equation}
where, $\Phi$ is the total power of the optical system, $\phi_i$ is the power of the $i^\text{th}$ individual lens, where there are $n$ lenses in the optical system.

The design of fisheye optics is an ongoing topic of research, with some recent advances in the use of neural networks for lens design \cite{tien2022design}. An optical system with more lenses does not necessarily imply a better-performing system. In one sense, a more efficient optical design implies using the optimum number of lenses. 
An eight-element wide FOV (120$^\circ$) lens was proposed in \cite{lee_2014_fisheye}. However, this system suffers from optical artifacts above 120$^\circ$ \cite{samy2015simplified}. In \cite{samy2015simplified}, a simplified fisheye design that optimizes optical artifacts is presented. Optical aberrations will be discussed in the following sections, where we will also survey the impact of the particular aberrations on the performance of computer vision.

\section{Fisheye Optical Aberrations}
\label{sec:optical-effects}
\begin{figure*}[t]
     \centering
     \begin{subfigure}{0.45\textwidth}
         \centering
         \includegraphics[width=0.9\textwidth]{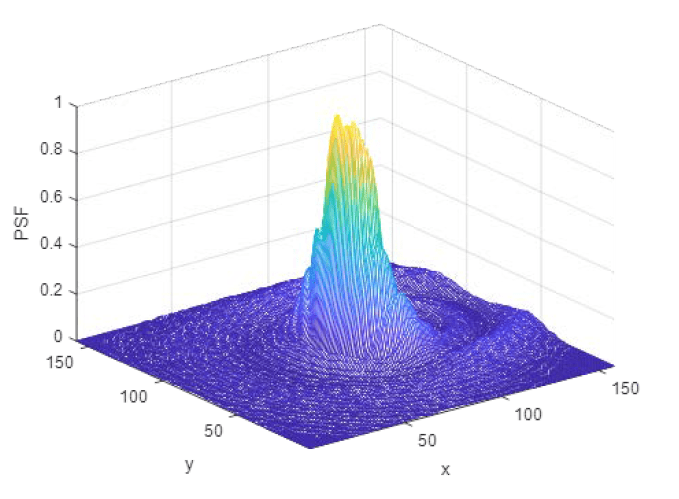}
         \caption{Centre}
         \label{fig:fisheye-PSF-centre}
     \end{subfigure}
     \begin{subfigure}{0.45\textwidth}
         \centering
         \includegraphics[width=0.9\textwidth]{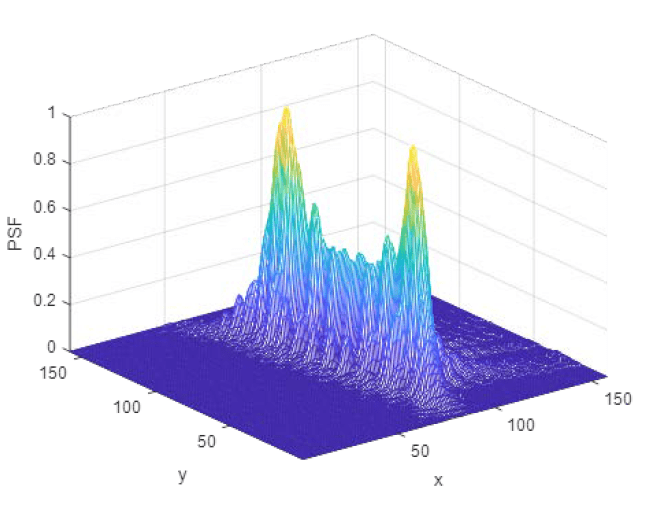}
         \caption{Edge}
         \label{fig:fisheye-PSF-edge}
     \end{subfigure}
        \caption{{\bf Measurement of PSF for a simulated fisheye lens at the center and edge locations} \cite{burns2020application}. Note that in the center, the PSF is approximately Gaussian, and has a relatively narrower peak indicating a sharper image. The Edge PSF is highly non-Gaussian and non-isotropic.}
        \label{fig:fisheye-PSF}
\end{figure*}

\begin{figure}[t]
    \centering
    \includegraphics[width=3.3in,clip,keepaspectratio]{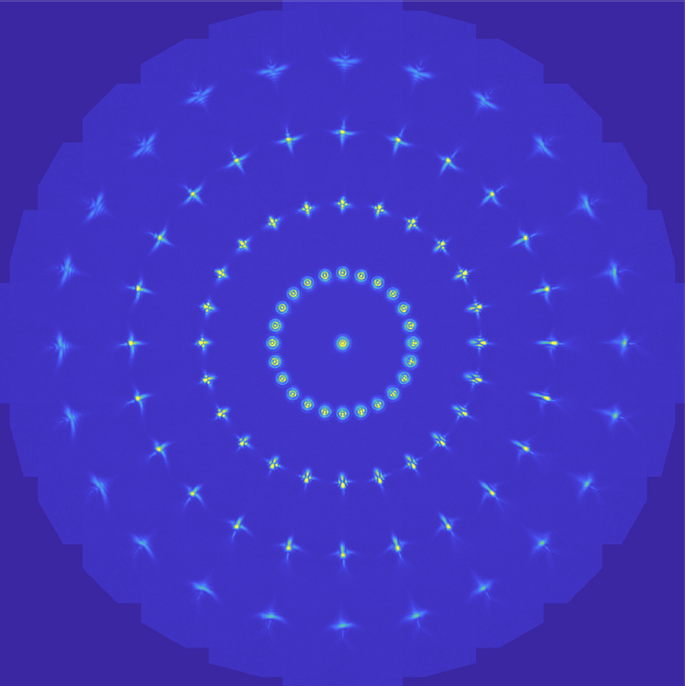}
    \caption{{\bf PSF model of a full image  \cite{wittpahl2018realistic, lehmann2019resolution}}. Note the highly varying individual PSFs, from approximately isotropic in the center to highly non-isotropic at the periphery.}
    \label{fig:psf-braun}
\end{figure}
In the previous section, we discussed the complexity of lens design and how aberrations are not possible to completely negate through optics. In this section, we look at what the major optical aberrations are and survey their impact on computer vision: (A) \ins{Spatially Variant Optical Quality}, (B) Chromatic Aberration, (C) Astigmatism, (D) Vignetting, and (D) Geometric Distortions. Specifically, the spatial variation on the image of the optical aberrations influences how objects are perceived by the computational observer.

\ins{\subsection{Spatially Variant Optical Quality}}

The spatially variant optical quality of a camera system can be measured by examining the image spots generated by rays propagating through the optical system (known as a spot diagram) \cite{krishnan2008cata}. The spots of light created by a camera on the image plane are known as the Point Spread Functions (PSFs) of a camera system and are essentially the impulse responses of the camera system. 
In normal optics, the PSF can be approximately Gaussian (e.g., Figure~\ref{fig:fisheye-PSF-centre}) \cite{yang2011adaptive, Eboli2022}. However, in fisheye optics, the PSF can be highly non-Gaussian (Figure~\ref{fig:fisheye-PSF-edge}). The spread of PSFs in the image is shown in Figure~\ref{fig:psf-braun}. 
Ideally, a camera system is spatially invariant where the same PSF describes the spatial impulse response of the camera at any point on the image sensor. However, fisheye cameras are spatially variant optical systems where optical quality can differ greatly across the image plane. 

\ins{Therefore, spatially variant optical quality should be considered when processing fisheye images. 
%Measuring optical quality has been a consistently evolving and debatable field in recent years especially now that artificial intelligence is being combined with cameras. 
While the measurement metrics for optical quality are of significant relevance, they will only be briefly discussed here and are not the main focus of this review. Traditional techniques exist for measuring the quality of cameras in the industry such as the Modulation Transfer Function (MTF) \cite{ISOBS2017, ISOBS2023} and signal-to-noise ratio (SNR) (peak SNR or PSNR is a studied metric for image quality assessment alongside MTF \cite{Muller_2023_ICCV, wang2004image}). Lin et al. \cite{lin2023simulating} considered evaluating image quality for motion blur and camera exposure based on both MTF and SNR techniques. However, the MTF has been considered insufficient for image applications in certain cases. For example, a study by Wolf et al. \cite{wolf2023windscreen} shows that MTF and Refractive Power metrics are not sufficient for measuring windscreen sharpness alone for artificial intelligence algorithms especially, where the camera is located behind the windscreen. As a result, other traditional metrics such as Shannon Information Capacity (SIC) \cite{koren2020measuring} and Noise Equivalent Quanta (NEQ) \cite{keelan2016imaging} are now being reconsidered as a way to evaluate computer vision performance and could potentially form part of a research direction in terms of computer vision for optical systems \cite{lin2023simulating}.}

As a result of the challenge of measuring and fine-tuning optical systems for artificial intelligence, recent work has investigated the impact of modeling and varying the optical parameters of narrow FOV optical designs using Neural Networks (NNs). In the work of Wittpahl et al. \cite{wittpahl2018realistic}, for a 60$^\circ$ standard FOV camera, an artificial neural network is trained to predict the spatially variant point spread function, given three input (1) defocus, (2) field and (3) azimuth. This can then be used to artificially model the impact of these three parameters on the optical quality of an image, whereby the PSF model is convolved with every pixel of a target image using PSFs determined by the ANN-based prediction, with downsampling and interpolation techniques using weighted kernels.

%As mentioned previously, PSFs are used to measure the degree of blurring in images created by an optical system. The PSFs illustrated in Figure~\ref{fig:fisheye-PSF-centre} - \ref{fig:fisheye-PSF-edge} can represent the way the system changes from the center to the edges or periphery of a camera. For fisheye cameras, the change in PSF reflects the change in the spatial domain of an optical system. A way of measuring PSF would be the Modulation Transfer Function (MTF) otherwise known as the Spatial Frequency Response (SFR) which should have a shape like that of a sinc function reflecting the impulse response of an optical system.

%There can be a wide range of PSFs scattered across an image which can look like an airy disk as illustrated in Wittpahl's work \cite{Wittpahl2018}. In Wittpahl's work, the \ang{60} FOV optical model was applied to example images, and areas of interest such as a car were cropped recreating a degraded image as if seen through the lens.

Furthermore, Lehmann et al. \cite{lehmann2019resolution} performed an investigation by examining the spatial resolution and accuracy of the model in \cite{wittpahl2018realistic}. The optimal topology of the ANN was found in this process by using Mean Square Error (MSE) as a decisive measure of modeling the 60$^\circ$ FOV model. It was found by a variation of between 8 to 448 neurons; the mean performance reached a plateau optimum of around 3 MSE within 50 to 100 neurons, with overfitting occurring when more neurons are added to the topology. For each topology configuration (i.e. variation in ANN) training was started 100 times with different randomized bias \cite{lehmann2019resolution}.

In principle, there is no reason that the models described in \cite{wittpahl2018realistic} and \cite{lehmann2019resolution} cannot be generalized to any optical model by design choice. For example, if a wide FOV camera model is to replace the narrow FOV model, in principle it is possible to reach a custom ANN topology for that specific model using the same approach and would be interesting future work. 

A caveat for the work of \cite{wittpahl2018realistic, lehmann2019resolution} should be highlighted. The images used were captured from a real camera. High-end optics were used in which the optical aberrations were small but non-zero. As such, the ANN-modeled PSF was applied on top of an existing PSF of the image lens, which will lead to distortion of the output image. As mentioned in \cite{lehmann2019resolution}, to apply the optical model to the pre-recorded data, the data taken by the original image lens would need to be deconvolved and then convolved with the optical model. A natural question arises as to whether it would be possible to apply the model directly to data collection in simulation environments (see Section~\ref{sec:sim}), which would be a fruitful direction of research.

\vspace{5mm}
\subsection{Chromatic Aberration}

Chromatic aberration occurs when long, medium, and short wavelengths of the visible spectrum (corresponding to red, green, and blue light) focus at different points in the image plane (i.e., do not converge at one specific point). This is an undesirable effect causing ``color fringes'' to appear at discontinuities (i.e., edges) in an image. %It causes difficulties further down the pipeline in camera production. 

\label{sec:ca-sys}
\begin{figure}[h]
     \centering
     \includegraphics[width=0.9\columnwidth]{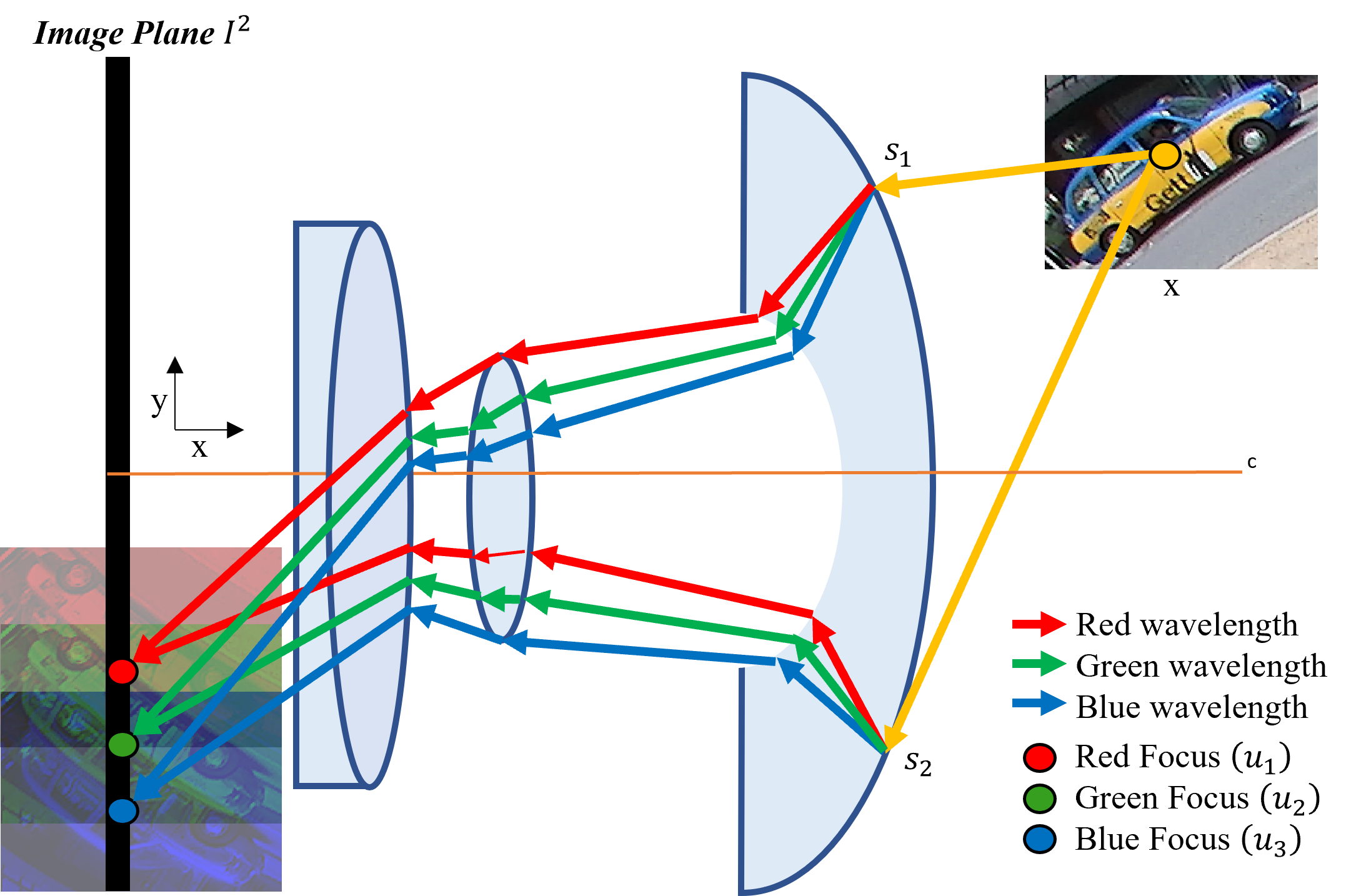}
     \caption{{\bf Lateral chromatic aberration on a fisheye lens}. Light splits into its components red, green, and blue (RGB) along the y-axis. We are only interested in the red, green, and blue components of the light, as an image sensor typically has only red, green, and blue pixels.}
     \label{fig:fisheye-lateral-CA}
\end{figure}

\begin{figure}[t]
    \centering
    \includegraphics[width=0.9\columnwidth]{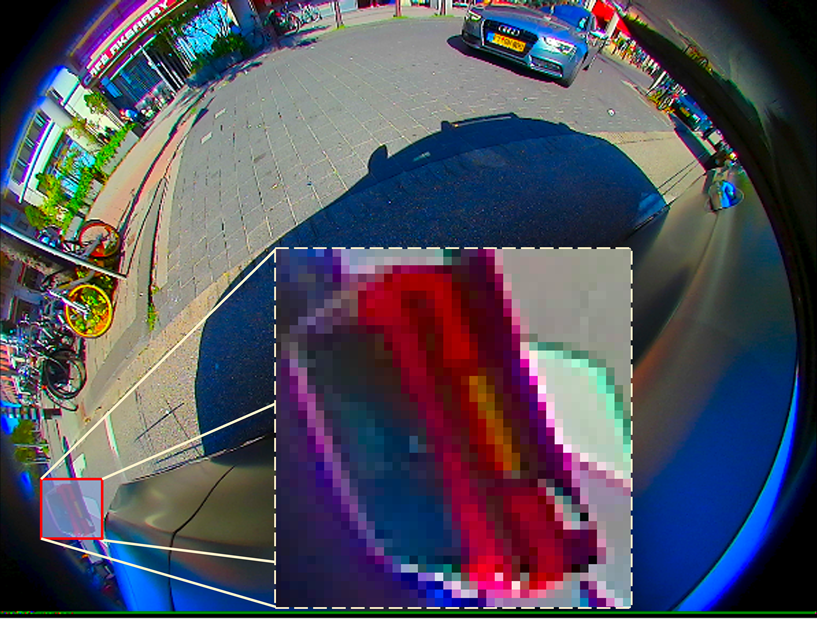}
    \caption{{\bf Zoomed-in segment of a Woodscape dataset image showing evidence of lateral chromatic aberration}. Note the slight red/purple and green tint in the pixels along the outline of the car and the road marking.}
    \label{fig:z_W}
\end{figure}

There are both lateral (i.e., light dispersing on the y-axis) and longitudinal (i.e., light dispersing on the x-axis) versions of chromatic aberration. Lateral chromatic aberration is demonstrated in Figure~\ref{fig:fisheye-lateral-CA}, where the further the rays of light fall from the center of the fisheye lens or optical axis, the wider the dispersion of light appears on the image plane. Lateral chromatic aberration typically increases towards the periphery of the FOV. As it is, essentially, a function of light ray incident angle, it is particularly noticeable and is very difficult to completely compensate for in fisheye lenses \cite{sun2017revisiting}. In Figure~\ref{fig:z_W}, it can be seen that lateral chromatic aberration exists in images in the fisheye Woodscape dataset \cite{yogamani2019woodscape}.

Conversely, longitudinal chromatic aberration occurs across the entire image regardless of the position from which light intersects the fisheye lens. In this case, the wavelengths of light which are longer (i.e., red and infrared) tend to converge past the image plane whereas shorter wavelengths (i.e., blue) converge before the image plane. Longitudinal chromatic aberration is easier to correct where stopping down a lens (i.e., increasing the f-number) compensates for this effect. Referring to Figure~\ref{fig:fisheye-lateral-CA}, there is no evidence of significantly visible longitudinal chromatic aberration in Woodscape.

Chromatic aberration is typically compensated for either through achromatic doublets in the optics (such as the optical doublet seen for the fisheye lens design in Figure~\ref{fig:fisheye}) or in the Image Signal Processing (ISP) stage before demosaicing of a camera system \cite{lluis2012chromatic,petersson2021blind}. Demosaicing is the stage at which a Bayer RAW signal (having one color per pixel) is converted into the standard RGB format. Images with lateral chromatic aberration perform poorly in demosaicing as the color effects shift the expected pixel detail in the resulting RGB images. Lateral chromatic aberration is usually corrected using Look-Up Tables calculated for a given lens type as a post-processing step, but this is often not feasible. Recent work has examined blind methods for correcting lateral chromatic aberration \cite{petersson2021blind}. It should be noted that the post-processing to compensate for chromatic aberration does not fix the problem. However, it is quite evident that it is not feasible to completely correct chromatic aberration in fisheye cameras.

Pontinen et al \cite{pontinen2012study} compared chromatic aberration between a Sigma 8mm 1:4 D EX Fisheye \cite{Sigma2007} and AF DX Fisheye-Nikkor 10.5mm f/2.8G ED \cite{Nikon2003} by applying different correction models to identify which is more suitable for 180$^\circ$ wide FOV lenses. The four correction models used in this study were shifting and scaling, affine, projective, and polynomial. These models were applied on the red and blue color channels where the green color channel was the reference channel for correction. It was found that the shifting and scaling correction model is adequate whereas the other models didn't improve chromatic aberration correction overall. What was interesting from this study was the analysis of the camera behaviors, where the change in aperture size and focus settings on the cameras influenced chromatic aberration unique to each camera design. For example, with increasing aperture size from f/4 to f/32, the red channel aberration increased by 0.5 pixels for the Sigma lens. With the Nikon lens, there was very little variation for the red channel.

It is intuitive to think that chromatic aberration will negatively impact the performance of computer vision tasks, as it is generally considered an unwanted optical artifact. However, this is not completely clear (and there is relatively little work examining this). According to a 2019 study by Chang et al \cite{chang2019deep}, it has been shown both qualitatively and quantitatively that images with chromatic aberration outperform `all-in-focus' images (i.e., images without any optical aberrations) both in Depth Estimation and 3D Object Detection applications using the NYU Depth v2 dataset \cite{silberman2012indoor} and a subset of the KITTI dataset \cite{geiger2013vision}. \ins{Table \ref{tab:results} gives a detailed insight into the effects of different optical artifacts according to literature. Chang et al. \cite{chang2019deep} is the only paper to date that gives evidence of a chromatic aberrated lens and a freeform lens performing better than the baseline or original data. Further evidence of this behaviour can be seen in recent work where robustness is significant against coma (a form of chromatic aberration) for a ResNeXt50 network \cite{Muller_2023_ICCV}. Additionally, Depth Estimation for defocused NYUv2 images performs better than the baseline (\emph{all-in-focus} images) for a D3-Net architecture \cite{carvalho2018deep} (see Table \ref{tab:results}). Whereas in all other literature concerning chromatic aberration a negative or degrading performance is observed for Object Detection and Instance Segmentation applications. However, since the optical designs used in these experiments have evidence of both chromatic aberration and astigmatism, it cannot be conclusively determined that the performance degradation is caused by any one artifact.} Also in recent work \cite{horn2022investigations, horn2021investigation}, longitudinal chromatic aberration was explicitly exploited for depth estimation on a bi-convex lens using the different spectral channels of unsharpness. The utility of chromatic aberration in tasks involving depth (e.g., depth estimation and 3D bounding box tasks) may be explained by the fact that the amount of chromatic aberration present is dependent on the depth of the imaged object, and this relationship can potentially be exploited. However, these experiments were performed on narrow FOV cameras. As discussed, chromatic aberration can be particularly severe in fisheye images, and no work (to the authors' knowledge) explicitly examines the impact of fisheye lateral chromatic aberration on computer vision performance. Additionally, it is not clear if other tasks, such as object detection or instance segmentation, are impacted negatively by the presence of significant chromatic aberration.
\begin{figure}[t]
    \centering
    \includegraphics[width=0.95\columnwidth]{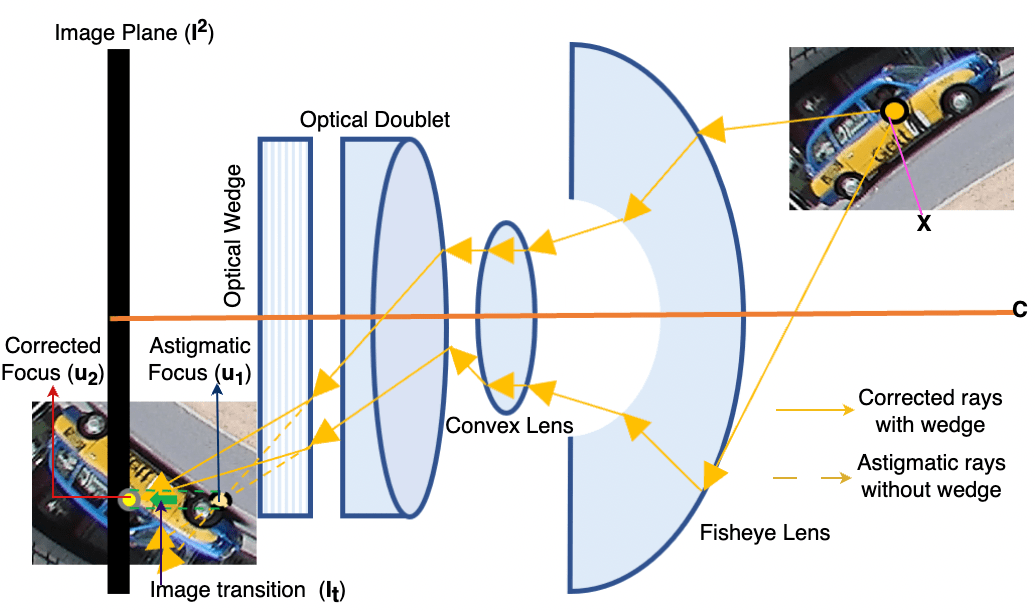}
    \caption{{\bf An optical wedge or prism used for astigmatism correction}. As shown in the diagram by adding the prism in front of an optical system; the rays are refracted to focus the light rays on the image plane producing a sharp image. }
    \label{fig:ast_c}
\end{figure}

\subsection{Astigmatism}
\label{sec:ast-sec}
\begin{figure}[t]
    \centering
    \includegraphics[width=0.9\columnwidth]{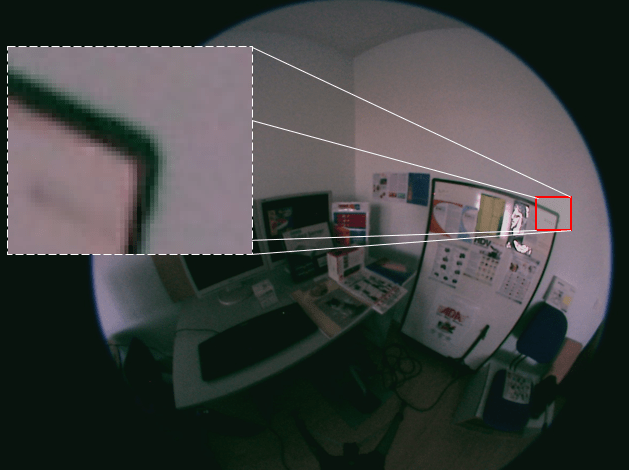}
    \caption{{\bf An example of a fisheye image displaying astigmatism}. Note the increased `blur' in the horizontal direction compared to the vertical. The Woodscape dataset does not exhibit significant astigmatism - this sample is taken from the non-automotive hybrid structure-from-motion and omnidirectional camera calibration dataset \cite{BASTANLAR2012557}. Note that some chromatic aberration and vignetting are also visible.}
    \label{fig:astigmatism_example}
\end{figure}

\begin{figure*}[t]
    \centering
    \includegraphics[width=0.9\textwidth]{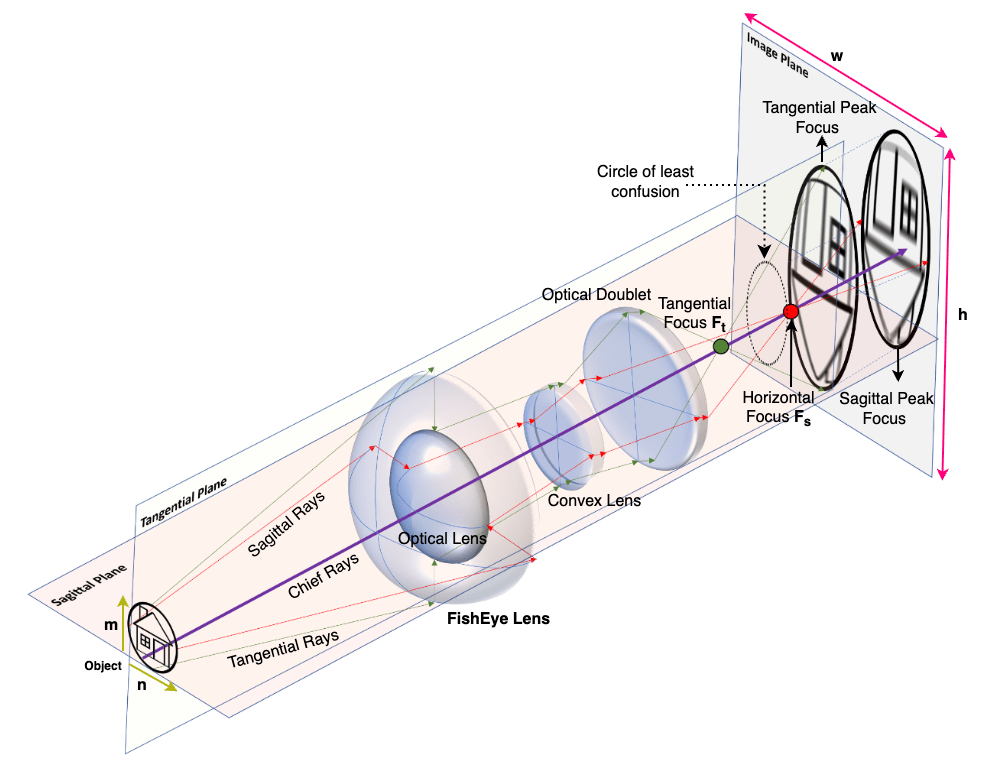}
    \caption{{\bf Demonstration of astigmatism in a fisheye lens}. Note the mismatching foci where the astigmatic images represent varying divergences of both sagittal and tangential rays.
    \label{fig:ast}}
\end{figure*}

Astigmatism is an optical artifact that impacts an image by causing a directionally dependent defocus (Figure~\ref{fig:astigmatism_example}). It is caused by an effect whereby rays in two perpendicular planes have two different foci, as shown in Figure~\ref{fig:ast}. It is introduced into the optical system either due to a lens not being rotationally symmetric or due to misalignment in lens elements and can degrade the performance of an optical system.
Concerning Figure~\ref{fig:ast}, the chief ray is a light ray passing through the center of the aperture of the system. From this ray, two different planes can be defined horizontally and vertically (also known as the sagittal and tangential planes). Likewise, there exist sagittal and tangential rays that form two unique foci (i.e., $F_{t}$, $F_{s}$). Hence two different images are formed (i.e., $I_{t}$, $I_{s}$). There can be many variations of sagittal and tangential planes by varying the angle of incidence of the chief ray \cite[274-276]{Hecht2017}. 
In an astigmatic system, it is impossible to focus both tangential and sagittal rays on one plane to create a sharp image. It has been noted that in an astigmatic system, the orientation of the object affects how imaging occurs \cite{deegan2015addressing}. For example, the greater the angle between the object and the optical axis, the greater the astigmatic difference is between $F_{t}$ and $F_{s}$, hence blurriness increases in the image.
Notice in Figure~\ref{fig:ast}, two elliptical images are formed where blurring is vertical for $I_{t}$ and horizontal for $I_{s}$. As astigmatism increases, the divergence between sagittal and tangential rays increases depending on manufacturing defects.  The \emph{circle of least confusion} is the smallest circle or spot that a lens system can make. The presence of astigmatism can change depending on lens manufacturing imperfections. A typical PSF in these conditions is shown in Section \ref{sec:fish-opt-sys} (Figure~\ref{fig:fisheye-PSF-edge}) (located at the periphery of the PSF lens model in Figure~\ref{fig:psf-braun}), which represents an astigmatic system occurring at the periphery of the lens where strong radial distortion is present. Astigmatism impacts the PSF of a given camera system directly (though it is not the only impact). There have been many attempts over the past few decades to correct astigmatism in optical systems. For example, the rear half of a classical symmetric Celor lens was used to reduce spherical aberration and astigmatism in a fisheye lens design \cite{geary2002introduction, samy2015simplified}. As described by Muller et al \cite{Müller2021}, the Cooke Triplet is an optical system with the ability to compensate for astigmatism. In that work, the Cooke Triplet was applied on a subset of the Berkely Deep Drive (BDD100k) \cite{bdd100k} automotive dataset for blurring between a defocus range of $\Delta{z}=\pm{1.25}$, where a Mask RCNN \cite{8237584} model was evaluated across this range of defocus. Within this range, it was shown that different Regions of Interest (ROI) from the center to the edge of an image have noticeably varying degrees of blurriness proving that the spatial domain of an image has inconsistent image quality and causes a significant drop in both statistical and spatial performance metrics for the Mask RCNN model. From these examples, it is clear that optical astigmatism has an undesired effect but is very difficult to control in complex optical systems.
For wide-angle lenses such as fisheye, astigmatism is particularly problematic where the radially varying PSFs (discussed in Section \ref{sec:fish-opt-sys}) affect the degree of blurring around the periphery of the lens.

Due to the mass production and time constraints of producing camera systems, this optical aberration cannot be eliminated. To compensate for astigmatism, optical systems could sometimes end up being rebuilt several times, but this is industrially undesirable  \cite{sasian2021control} \cite{brummel2021performance}.
To correct linear astigmatism, a manufactured lens can be rolled on its mechanical seat so that one surface becomes tilted by an angle of $\alpha$ measured in arc minutes which can be estimated by (\ref{eq:tilt}) \cite{sasian2021control}:
\begin{equation} \label{eq:tilt}
\alpha = \frac{d}{R}
\end{equation}
Where $d$ is the lateral displacement of the lens due to rolling and $R$ is the radius of curvature of the surface on the seat.
Finally, a thin optical wedge is added to the optical system just before the light reaches the image plane as described by\cite{sasian2021control,sasian2000aberrations}:
\begin{equation} \label{eq:wla}
W_{LA} = -\frac{1-n^2}{n}*u*\bar{u}*\alpha*y
\end{equation}
Where $n$ is the index of refraction of the wedge, $u$ is the slope of the first order marginal ray before entering the wedge, $\bar{u}$ is the slope of the first order chief ray before entering the wedge in image space respectively, and $y$ is the marginal ray height at the wedge \cite{sasian2021control}. Figure~\ref{fig:ast_c} demonstrates these parameters on a sample optical wedge-shaped as a rectangular prism.

It has been demonstrated that astigmatism can be undesirable for computer vision algorithms as it can influence the performance of object detection and segmentation algorithms due to the blurriness of the target objects \cite[p. 43]{brummel2021performance}. \ins{Substantial evidence in literature all demonstrate this for computer vision \cite{Müller2021, carvalho2018deep, brummel2021performance, Muller2022, brummel2022spatial} (see Table \ref{tab:results})}. An example of a negative impact on performance can be seen in \cite[p. 60-61]{Müller2021}, where it is demonstrated that using a Mask-RCNN model the Average Precision (AP) worsens in performance on the Precision-Recall (PR) curves by applying a defocus of $\Delta{z}=\pm{1.25}$ using the Cooke Triplet lens. Contrarily, a positive impact can be seen in 3D Object Detection results from Chang et al. \cite{chang2019deep}, using an optimized mask for a freeform lens initialized with astigmatism. For example, when trained on 5500 KITTI \cite{geiger2013vision} images the optimized lens model produced 37.51\% Average Precision (AP) which is 10.8\% higher for easy 3D object localization when compared to the all-in-focus (original) KITTI dataset. Similarly, there is an 8.34\% AP improvement for `easy' 3D object detection (`easy' being a difficulty parameter of the KITTI performance metrics). When considering astigmatism for the application of Depth Estimation, a popular term was coined in the 20$^\text{th}$ century called Depth-from-Defocus (DfD) \cite{subbarao1994depth}. Using DfD, Carvalho et al. \cite{carvalho2018deep} synthetically defocused datasets (such as NYUv2 \cite{silberman2012indoor}) and depth was learned from the defocused blur showing a performance improvement when compared to `all-in-focus' (unaugmented) images. As with the PSF, machine learning tasks that require depth estimation may benefit from the depth-dependent defocus.
\subsection{Vignetting}
Vignetting (otherwise known as `light fall off') is an optical phenomenon that causes gradual darkening at the periphery of an image. Vignetting occurs due to the varying angle of light across a sensor, the intrinsic lens characteristics, and physical occlusions \cite{tian2021ava}. It is most apparent in extremely wide FOV cameras, such as fisheye. There are four main causes of vignetting \cite{potovc2022creating}: %: (1) natural, (2) optical, (3) pixel and (4) mechanical.
\begin{enumerate}
    \item \textbf{Mechanical vignetting} is the physical occlusion of light to a camera sensor which gives a loss of information in the resulting image \cite{potovc2022creating}. This is generally a trade-off in the design of fisheye cameras. Reducing the mechanical vignetting would necessitate the sacrifice of camera FOV, with most fisheye cameras designed either to be full-circle (ref Figure~\ref{fig:astigmatism_example}) or cropped-circle (ref Figure~\ref{fig:z_W}).
   \item The \textbf{cosine-fourth-power-law} describes the illumination fall off due to the apparent elliptic eccentricity of the entrance pupil from the point of view of the incident light ray \cite{kerr2007derivation}. The greater the off-axis angle of the object being imaged the more elliptic the entrance pupil becomes. This increased eccentricity of the entrance pupil from the point of view of the object means less light propagates through the pupil, and hence we have an illumination fall off.
   \item \textbf{Optical vignetting} occurs due to physical occlusion between lens elements (literally, rear elements are shaded by those in front). It can often be reduced by reducing the diameter of the aperture (otherwise known as `stopping down the lens') \cite{samy2015simplified}. 
   \item \textbf{Pixel vignetting} is caused by the physical structure of the pixels in an image sensor \cite{10.1117/12.385460}. In an image sensor, the top layer of the sensor is not (typically) the light-sensitive layer. When light is incident on a pixel, it must travel through a `tunnel' (layers of CMOS) before it hits the photodiode. The deeper this `tunnel' is relative to the pixel area, the more a shadow is cast by the pixel walls, causing increased light fall off with increased incident angle.
\end{enumerate}
\begin{figure}[t]
    \centering
    \includegraphics[width=0.8\columnwidth]{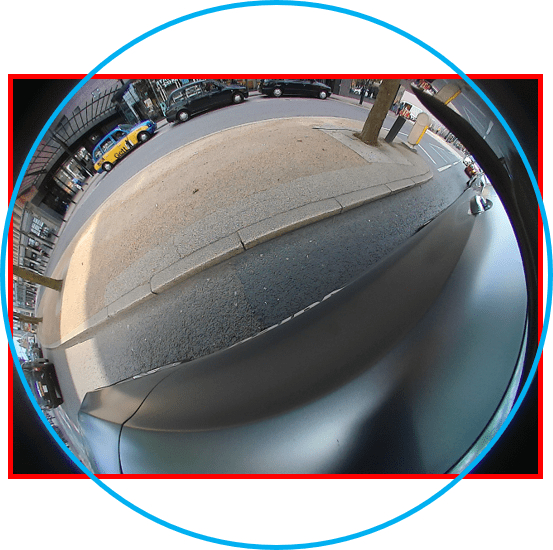}
    \caption{{\bf Mechanical vignetting in the Woodscape dataset}. Note the presence of black corners due to a mismatch in the sensor and lens aperture.}
    \label{fig:mech-vig}
\end{figure}
Both optical and pixel vignetting can be compensated for at the image signal processing stage (ISP) of a camera system \cite{potovc2022creating, jimaging5100078}.
For the Woodscape dataset, vignetting is largely compensated with radial digital gain correction using a lens shading block \cite{yogamani2019woodscape}. This post-processing compensation may fix the visual appearance of the vignetting, but the signal-to-noise ratio will not be improved. There is also mechanical vignetting along the aperture, as seen in Figure~\ref{fig:z_W}. The ``black corners'' that are distinctly visible are due to the lens hood that prevents the image sensor from seeing the complete FOV. The on-automotive hybrid structure-from-motion and omnidirectional camera calibration dataset \cite{BASTANLAR2012557} shows some level of optical/pixel/cosine-fourth vignetting (ref Figure~\ref{fig:astigmatism_example}).

The ISO12233 standard \cite{iso112233} has introduced polynomial fitting and non-uniform compensation to correct for vignetting and geometric distortion effects in Edge Spatial Frequency Response (e-SFR) measurements \cite{burns2022updated}. When high (1 side) and additional low-signal (2 sides) corrections are applied, the SFR decreases in both stages, hence reducing vignetting in measurements. ISO12233 is used in measurements for surround-view cameras in autonomous vehicles \cite{parulski2022creation}. In terms of computer vision performance, yet again, somewhat counterintuitively, vignetting has been shown to have some positive impact on statistical metrics for object detection. For instance, it has been shown that applying vignetting to the KITTI \cite{geiger2013vision} and Virtual KITTI \cite{gaidon2016virtual} datasets for 2D object detection on cars \cite{saad2019camera}, where a Faster-RCNN \cite{renNIPS15fasterrcnn} model with a region proposal network (RPN) (pre-trained on ImageNet \cite{deng2009imagenet}) was used, precision increased from 83.12\% to 88.96\% when vignetting was applied to the images \cite{saad2019camera} (see Table \ref{tab:results}). Also, the number of car detections increased at the periphery of the images, which is unusual as one would assume that vignetting would cause a performance drop at the periphery.
However, precision alone is not sufficient for completely determining the apparent positive impact of vignetting on performance, as it is well-known that precision is biased towards the positive class \cite{powers2020evaluation}. As a result of insufficient statistical metrics in these experiments, more investigation is needed on the impact of vignetting on computer vision performance.
On the contrary, an Adversarial Vignetting Attack (AVA) model confused a ResNet50 \cite{he2016deep} network which demonstrates that vignetting can have an unpredictable effect on computer vision performance \cite{tian2021ava} (see Table \ref{tab:results}). More experiments are needed to understand these behaviors. \ins{Table \ref{tab:results} provides a detailed comparison of experiments in the literature aimed at assessing optical artifacts in computer vision. It is reasonable to deduce that, among the identified papers, both negative and positive effects of optical artifacts can occur depending on the nature of the examined tasks. It cannot be conclusively assumed that introducing optical artifacts into training data results in degraded performance. For example, Depth Estimation experiments show a positive impact, and augmenting synthetic data improves performance on real-world data \cite{carlson2018modeling}. It is also evident that individual metrics, such as precision, improve when vignetting is applied to KITTI (see Table \ref{tab:results}). However, degrading real-world images with vignetting as an adversarial attack on algorithms gives degrading performance \cite{tian2021ava}. This demonstrates that image degradation can lead to positive results for datasets with simulation but negative on real-world data.}
\begin{table*}
\vspace{-0.415in}
\ins{
\scriptsize
\centering
\caption{Comparison of Optical Artifacts on Computer Vision}
\label{tab:results}
\begin{threeparttable}
\resizebox{\textwidth}{!}{%
\begin{tabularx}{7in}{|l|l|l|l|l|l|p{2.4917in}|}
\hline
\multirow{2}{*}{\textbf{Category}} &
  \multirow{2}{*}{\textbf{Year}} &
  \multirow{2}{*}{\textbf{Author}} &
  \multirow{2}{*}{\textbf{Dataset}} &
  \multirow{2}{*}{\textbf{Tasks}} &
  \multirow{2}{*}{\textbf{Impact}} &
  \multirow{2}{*}{\textbf{Computer Vision Performance}}
  \\
  \multicolumn{1}{|c|}{} &
  \multicolumn{1}{c|}{} &
  \multicolumn{1}{c|}{} &
  \multicolumn{1}{c|}{} &
  \multicolumn{1}{c|}{} &
  \multicolumn{1}{c|}{} &
  \\ \hline
  \multirowcell{6}{Chromatic Aberration\\\& Astigmatism}&
  2018 &
  \cite{carvalho2018deep} &
    \multirowcell{1}{NYU Depth v2} &
   Depth Estimation &
   \cellcolor{green!25}positive &
 Using the D3-Net architecture for the Depth from Defocus approach proposed outperforms `all-in-focus' for NYU Depth v2  RGB images with additional blur where root mean squared error (RMSE) improved from 0.706 to 0.144 which is about 79.6\% error reduction. \\ \cline{2-7}
  &
  2019 &
  \cite{chang2019deep} &
    \multirowcell{2}{KITTI,\\NYU Depth v2,} &
  \multirowcell{2}[0pt][l]{3D Object Detection,\\Depth Estimation,}&
  \cellcolor{green!25}positive &
  The second lowest linear MSE for Chromatic Aberration of 0.4496 (NYU Depth v2), and 2.2566 (KITTI). The third lowest linear MSE for Astigmatism of 0.4561(NYU Depth v2) and 2.3634 (KITTI). Significant improvements can also be seen for 3D Object Detection where the freeform lens improved by +8.34\% AP(Easy mode), +3.2\% AP (moderate mode) and +0.17\% AP (hard mode) respectively. Perhaps surprisingly, this work demonstrates that a simple lens with chromatic aberration performs almost as well as a freeform lens for the tasks discussed. There is, however, some further work that backs this up \cite{jung2021compact}.
  \\ \cline{2-7}
  &
  2021 &
  \cite{brummel2021performance} &
    \multirowcell{2}{BDD100k,\\Cityscapes,} &
   \multirowcell{2}[0pt][l]{Object Detection,\\Instance Segmentation,} &
   \cellcolor{red!25}negative &
  Cooke Triplet model applied to datasets for a defocus range between $\Delta$z = \{-1.25, -0.75, -0.5, 0, +0.5, +0.75, +1.25\}. For car detection, the baseline car 78\% AP degraded to between 67\%-76\% AP for BDD100k. For car instance segmentation, a baseline of 68.9\% AP degraded to between 58.9\%-65.69\% AP for Cityscapes \\ \cline{2-7}  
&
  2021 &
  \cite{Müller2021} &
     \multirowcell{1}{BDD100k}&
   Object Detection, &
   \cellcolor{red!25}negative &
  Applying defocus parameters to BDD100k images shows a clear drop in spatial performance of 7\% AP at the periphery of the spatial domain for $\Delta$z = -1.25 and 5.4\% AP for $\Delta$z = +1.25. Overall results for degraded BDD100k show an improvement from a baseline of 90.72\% AP to between 83.758\% and 89.24\% AP. \\ \cline{2-7}
   &
  2022 &
  \cite{Muller2022} &
     \multirowcell{1}{BDD100k} &
   Object Detection &
   \cellcolor{red!25}negative &
  10k images of BDD100k for the baseline (original images) 61\% AP degraded by a Cooke Triplet model to between 54\%-55\% AP where little variations in results were observed for approximation models. \\ \cline{2-7} 
  &
  2022 &
  \cite{brummel2022spatial} &
    \multirowcell{1}{BDD100k,} &
   Instance Segmentation &
   \cellcolor{red!25}negative &
  Cooke Triplet model was applied to BDD100k from \cite{Müller2021}. SPI and SRI metrics LAMR* increased from 0.6367 (baseline) to between 0.64361 and 0.68354 (degraded). \\ \cline{2-7}
  &
  2024 &
  \cite{molloy2024analysis} &
    \multirowcell{2}{Primary Research} &
   \multirowcell{2}[0pt][l]{Object Detection} &
   \cellcolor{green!25}positive &
  The first major comprehensive analysis of primary research data for pedestrians and cars collected using a BlackFly-S 8.9MP camera and degraded between $\Delta$z = \{-1.5, -1.0, -0.5, 0, +0.5, +0.1, +1.5\}. Results show for a degraded dataset of $-0.5\lambda$, object detection performance for Average Precision between a 50\% and 95\% IoU threshold was in most cases higher than the baseline across six different deep learning models. Out of all six models Faster RCNN Swin Tiny FPN was the most robust against degradation achieving a performance difference of 7.22\% (large objects) and 20.72\% (small objects).\\ \hline  
  Chromatic Aberration &
  2018 &
  \cite{carlson2018modeling} &
  \multirowcell{1}{KITTI,} &
  Object Detection &
  \cellcolor{yellow!25}neutral &
   Modeling optical artifacts by degrading synthetic data (i.e., VKITTI and GTA) to improve performance on real-life data (i.e., KITTI). For 21k VKITTI images, 58.25\% AP (baseline KITTI) improved to 62.52\% (KITTI with degraded VKITTI). Similarly, for 50k GTA, 53.26\% AP (baseline KITTI) improved to 55.85\% AP (KITTI with degraded GTA) on the same held-out set of KITTI. While introducing Chromatic Aberration into simulation shows improvements; it does not prove it has either a positive or negative impact on real-life data.\\ \hline
Vignetting &
  2019 &
  \cite{saad2019camera} &
    \multirowcell{1}{KITTI,} &
   Object Detection &
   \cellcolor{green!25}positive &
   Applying vignetting to KITTI images using an integration sphere improved precision from 83.12\% to 88.96\% which (somewhat surprisingly) gives a gain of 5.84\% for the Faster RCNN with ResNet50. \\ \cline{2-7} 
  &
  2021 &
  \cite{tian2021ava} &
    \multirowcell{1}{DEV,\\CIFAR-10,\\Tiny ImageNet,} &
  \multirowcell{2}[0pt][l]{Adversarial Attack \&\\Classification} &
  \cellcolor{red!25}negative &
  Radial Anisotropic AVA (RA-AVA) degraded images from DEV, CIFAR-10, and Tiny ImageNet datasets producing either the highest or second highest transfer success rates for all. Neural Network architectures such as  ResNet50 gives degraded performance of 46.34\% compared to the baseline (original) for Tiny ImageNet. \\ \hline
  Distortion &
  2019 &
  \cite{deng2019restricted} &
  \multirowcell{3}{Cityscapes,\\SynWoodscape,\\SVScape} &
  Semantic Segmentation &
  \cellcolor{red!25}negative &
   Restricted deformable convolution improved the performance on real-world data to as high as 74.2\% mIoU using the ERFNET-RDC-8 which is a VGG-based architecture neural network \cite{simonyan2014very}. Proposing a new architecture for distortion gives an alternative over the traditional Computer Vision algorithms and is likely the only way to remove performance degradation of this optical artifact.\\ \cline{2-7}&
  2023 &
  \cite{shen2023optical} &
  \multirowcell{2}{Woodscape,\\SynWoodscape,} &
  Optical Flow &
  \cellcolor{red!25}negative &
   RAFT network is finetuned using SynWoodscape and gives the lowest End Point Error (EPE) on Woodscape (1.40) compared to both Sintel (5.12) and Jointly trained (1.48) versions. However, results signify that Distortion does not have a good effect on traditional optical flow.\\ \cline{2-7}   
 &
  2023 &
  \cite{cho2023surround} &
  \multirowcell{1}{Woodscape,} &
  Semantic Segmentation &
  \cellcolor{red!25}negative &
   Viewpoint Augmentation gives the highest result (66.91\% mIoU) compared to both images without augmentation (63.80\% mIoU) and with affine augmentation (66.60\% mIoU). Using augmentation to improve results further shows evidence that traditional Computer Vision algorithms are not able to handle distortion well\\ \hline
\end{tabularx}
}
\vspace{0.1in}
\begin{tablenotes}[para,flushleft]
\centering
  Note: The Impact column signifies any quantitative difference in statistical metrics considering optical artifacts where a positive impact implies a better performance in metrics which includes but is not limited to Average Precision (\% AP), Mean Squared Error (MSE), Precision, End-point Error (EPE), mean Intersection over Union (mIoU).
\end{tablenotes}
\end{threeparttable}
}
\end{table*}
\subsection{Geometric Distortion}
\label{sec:geo-met}
The geometry of fisheye cameras has been covered in several previous surveys \cite{kumar2023surround, hughes2009}, and as such we only give a brief discussion here on the topic.

The very strong radial distortion in fisheye optics is, perhaps, the most obvious undesirable artifact.
In \cite{kumar2023surround}, there is an in-depth discussion on different fisheye mathematical models.
It is especially interesting to note a fisheye camera's relationship with the 2D image of size $I^2$ from a geometrical perspective. Here we give a very brief overview but would encourage readers to examine the previous surveys for more details.
\begin{table*}[]
\ins{
\scriptsize
\centering
\caption{Optical Lens Modelling with Artificial Intelligence}
\label{tab:ai-lens-model}
\begin{threeparttable}
\resizebox{\textwidth}{!}{%
\begin{tabularx}{7in}{|l|l|l|l|p{3.41in}|}
\hline
  \multirow{1}{*}{\textbf{Year}} &
  \multirow{1}{*}{\textbf{Paper}} &
  \multirow{1}{*}{\textbf{FOV}} &
  \multirow{1}{*}{\textbf{Type of Model}} &
  \multirow{1}{*}{\textbf{Computer Vision Performance}}
  \\ \hline
  2017 &
  \cite{zheng2017neurolens} &
    160º &
   Multi-Layer Perceptron &
  Neurolens first proposed and simulated for four different lenses including fisheye where an acyclic four-layer perceptron network is trained ((1x) input layer, (2x)hidden layers and (1x) output layer). The number of layers is increased if the prediction and testing error ($\epsilon$) exceeds (0.2\% - 0.8\%). \\ \hline 
  2018 &
 \cite{wittpahl2018realistic, lehmann2018resolution, lehmann2019resolution} &
    60º &
    Neural Network &
  The first proposal of realistic image degradation using a spatially variant distribution of the PSF of an optical lens trained using non-linear regression with an ANN. Results show that training the model with the three input parameters defocus ($\Delta$z), image height (R) and azimuth ($\phi$) gives a qualitatively realistic degradation. Lehmann et al. \cite{lehmann2018resolution, lehmann2019resolution} optimizes the same model with Mean Squared Error (MSE) reaching an optimal of 3 within 50 to 100 neurons. \\ \hline
  2023 &
  \cite{xian2023neural} &
     \multirowcell{1}[0pt][l]{OC*(130º - 266º),\\UZH \cite{delmerico2019we}(124º - 166º)} &
   Neural Network &
  A second Neurolens is proposed for distortion and vignetting for the Lensfun database and is extended to OC and UZH generalizing well across different lenses. \\ \hline 
  2023 &
  \cite{yang2023curriculum} &
    68.8º &
   \multirowcell{2}[0pt][l]{Curriculum Learning\\(Reinforcement Learning)} &
 Autolens is proposed in this work where Curriculum Learning is used to design a pinhole camera model based on geometrical optics and is compatible with Zemax OpticStudio. \\ \hline 
  2023 &
  \cite{yang2023image} &
    68.8º &
   Neural Network &
 Task-Driven Lens modeling from scratch uses supervised lens design to outperform conventional lens systems in image classification where for an optical doublet ResNet50 classification accuracy improved by 1.54\% for the TaskLens model and for a Large Patch vision transformer (ViT-L/16) by 0.57\% respectively. \\ \hline
\end{tabularx}%
}
\vspace{0.1in}
\begin{tablenotes}[para,flushleft]
\centering
  Note: OC* = Omnidirectional Camera Calibration (OCamCalib), DGauss = Double Gaussian lens
\end{tablenotes}
\end{threeparttable}
}
\end{table*}
\begin{figure}[h]
    \centering
    \includegraphics[width=0.9\columnwidth]{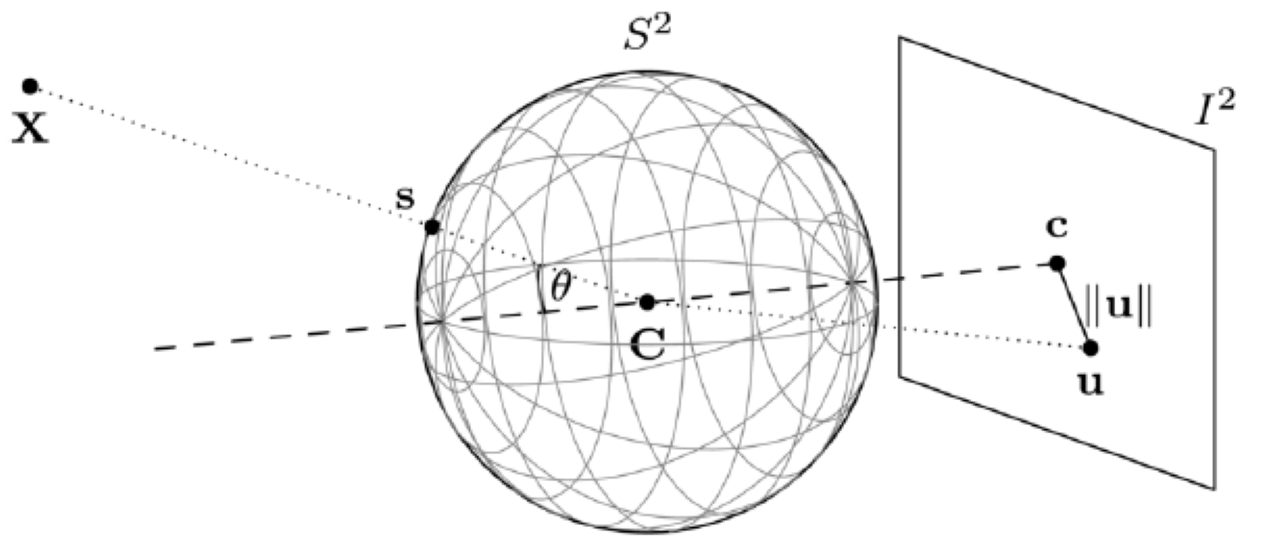}
    \caption{{\bf Geometric Relationship between a fisheye point at $u$ on the two-dimensional image and an equivalent point at $s$ on a unit sphere} \cite{kumar2023surround}. The projective sphere is a more natural geometric object for modeling the rays of an omnidirectional/fisheye camera, in the same manner that the projective plane is used for narrower FOV cameras \cite{Mariotti2022}.}
    \label{fig:geom-sv}
\end{figure}
In Figure~\ref{fig:geom-sv}, $\theta$ is the angle at which the ray of light from an object X maps from a unit sphere onto an image plane. $C$ is the center of the unit sphere. The radial distance $||\textbf{u}||$ of the projected image point $\textbf{u}$ to the image center $\textbf{c}$ is proportional to this incident angle $\theta$. Some of the classic geometric projection models that represent fisheye distortion include (1) equidistant, (2) stereographic, and (3) orthographic projections. Each projection model has a slight variation of the general mapping strategy illustrated by Figure~\ref{fig:geom-sv} \cite{kumar2023surround}. Each model has its own set of unique functions for both projection and unprojection and these models in theory can be applied to fisheye images to either remove the distortion or add distortion to rectilinear images. The polynomial Brown-Conrady model (including calibration methods), is implemented in many commonly used software packages and libraries (e.g., MATLAB \cite{MATLABundist2022, MATLABCameraCalib2022}).
The popular Woodscape datasets \cite{yogamani2019woodscape, Sekkat2022}, examined closely in this paper, use a full fourth-order polynomial model for fisheye projection:
\begin{equation} \label{eq:poly}
\|\textbf{u}\| = r\left(\theta\right)=a_1\theta+\ a_2\theta^2+\ a_3\theta^3+\ a_4\theta^4
\end{equation}

\ins{Neural networks are not specifically built for distorted images which lead to an expected degrading performance \cite{eising2021near}. For example, due to the distortion in fisheye imagery, the implicit assumption of translation invariance in CNNs does not hold true. Furthermore, the use of augmentation methods does not necessarily imply more data equates to better performance. Cho et al. \cite{cho2023surround} present viewpoint augmentation as the method that gives the least degrading performance for fisheye (see Table \ref{tab:results}). %However, it is possible to finetune computer vision algorithms via transfer learning as shown in Section \ref{sec:compvis} and Figure \ref{fig:yolo-mvl} which gives an overall higher IoU and fewer false positives (FPs) than training from scratch on distorted images.
%A notable observation from the literature shows that synthetic distorted images can be used as a strategy for improving computer vision performance (e.g., optical flow when combining Woodscape with SynWoodscape \cite{shen2023optical}(see Table \ref{tab:results})).
More recent developments show that deformable convolution networks \cite{dai2017deformable} would effectively solve the distortion problem when considering the use of offsets and geometric transformations in images. Deng et al. \cite{deng2019restricted} propose a restricted version of this strategy where the model was shown to be robust for large radial distortions in surround-view systems (see Table \ref{tab:results}).
}

\subsection{Discussion}

What is perhaps of most interest is that, when surveying the admittedly sparse literature, it is not clear that all optical aberrations impact the performance of all automotive computer vision tasks negatively (see Table \ref{tab:results}). Several pieces of work have demonstrated that certain tasks have been improved by optical aberrations. For certain tasks, such as those that extract depth from the image, this can make some sense due to the depth-dependent nature of certain optical aberrations. Considering the sparsity of the literature, it is evident that further work in this space is required. 

It should be noted that there are some elements related to optics that we haven't discussed completely in this section. For example, images taken by cameras are also affected by natural elements of the scene, such as sunlight causing sun flare (as is the case in Fig \ref{fig:fisheye}). 
False artifacts appear near the aperture of the lens in scenes with sun flare \cite{jakubiec2016accurate}. This produces unwanted blurs at the periphery of the image and can be mistaken by computer vision systems when performing object detection. Actors such as pedestrians and cars can be occluded by the sun flare, leading to missed detections. This is especially critical when the sun is low in the sky in which case sun flare can be unavoidable despite efforts to remove it in camera production.

Additionally, there is some work ongoing on designing optical systems for computer vision. Designing optical lenses for automotive computer vision tasks is especially challenging given the requirement of using high-quality and standard high-resolution input images for the training of DNNs and other recent approaches (e.g., Vision Transformers \cite{dosovitskiy2020image} and Bidirectional Encoder representation from Image Transformers \cite{bao2021beit}). Research on lens design optimization concerning computer vision is ongoing. In Yang et al \cite{yang2023image}, a Task-Driven End-to-End Lens design approach has been introduced as a more feasible option over conventional lens designs. End-to-end optical design is a relatively new area of research that has shown promising results when applied to common computer vision tasks, including object detection. However, these lens designs rely on well-developed End-to-End optical design solutions \cite{yang2023image, sun2021end, li2021end, tseng2021differentiable}. However, for example, the TaskLens model \cite{yang2023image} shows a lot of potential and unexplored areas in optical design outperforming the image classification of three traditional lens designs developed using conventional lens design methods (i.e., the doublet, triplet, and quadruplet each having a 68.8$^\circ$ FOV and f-number of f/2.8). \ins{There is some very recent work showing a trend in the adoption of optical lens modeling using artificial intelligence as a strategy to design for computer vision performance as shown in Table \ref{tab:ai-lens-model}}.

\section{Computer Vision in Surround-View Cameras}
\label{sec:compvis}
In the previous section, we briefly discussed the impact of individual optical artifacts. In this section, we examine computer vision performance in surround-view cameras as a whole. This is illustrated using an application of the YOLOv7 object detection algorithm on the Woodscape dataset. 
What is evident from the previous section is that for fisheye, the optical artifacts are strongly spatially variant (in narrow FOV cameras, this spatial variance can typically be safely ignored). Therefore, perhaps, global metrics of performance do not give us the full picture of the performance of computer vision from fisheye cameras. In this section, we therefore also briefly discuss recent proposals on spatially dependent performance metrics.

\subsection{Computer Vision Performance}
\label{sec:cv-fisheye}
\begin{figure*}[t]
     \centering
     \includegraphics[width=7in, keepaspectratio]{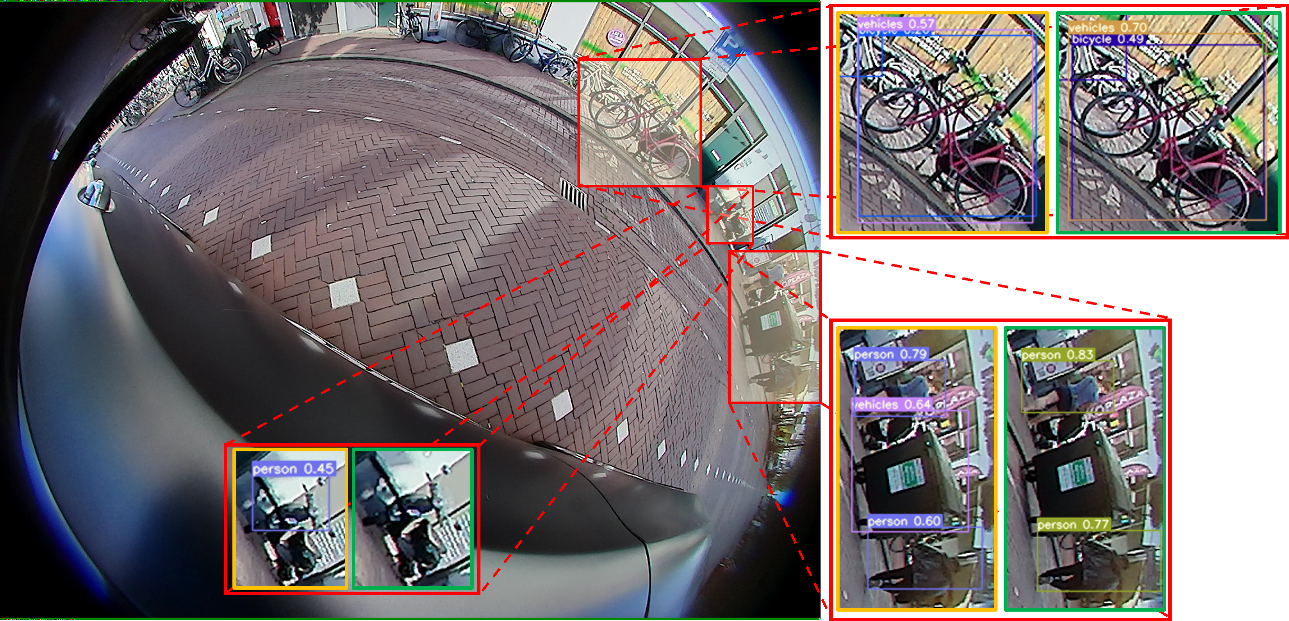}
    \caption{\ins{{\bf YOLOv7 inference regions of interest (ROIs) (red box) on the Woodscape left camera image was performed. Qualitative results show both training from scratch (orange box) and transfer learning (green box)}. Notice the misclassification of object types in both instances. The orientation of the object in the image due to its location can confound an object detection network. Note: the pre-trained model used in transfer learning was pre-trained on the MS COCO dataset.}}
    \label{fig:yolo-mvl}
\end{figure*}
Computer Vision research is a consistently developing field where new algorithms are being developed or optimized to solve the perception problem in images. To get an idea of the speed of development, You-Only-Look-Once version 8 (YOLOv8) \cite{Jocher_YOLO_by_Ultralytics_2023} was released approximately 5 months after the previous version, i.e., YOLOv7 \cite{wang2023yolov7}. To give an indication of the rate performance improvement for YOLO, using the baseline models comparison, YOLOv7, when trained on the MS COCO \cite{lin2014microsoft} dataset with 75\% fewer parameters than YOLOv4 \cite{bochkovskiy2020yolov4}, is capable of reaching 66.7\% AP which is 1.5\% better than YOLOv4 \cite{wang2023yolov7}.
The general trend for each YOLO version release is based on performance optimization such as speeding up non-maximum suppression (NMS) and maintaining previous structural improvements such as being an anchor-free network \cite{terven2023comprehensive}.

As we have already seen, despite the highly dynamic nature of computer vision research at the moment, there is relatively little research done into how optical characteristics, discussed in Section \ref{sec:optical-effects}, affect computer vision statistical metrics. As discussed by Brummel et al. \cite{brummel2022spatial}, spatial metrics are a more insightful approach than statistical metrics (we shall discuss these in the next subsection). To our knowledge, to date, there hasn't been a more rigorous analysis comparing both metrics across the FOV of fisheye. This remains an open challenge for the community. However, here we will give some indication of the complexity of the problem.

Figure~\ref{fig:yolo-mvl} shows the output of YOLOv7 inference run on a WoodScape sample. \ins{In the Regions of Interest (ROIs) chosen from the image both training from scratch and transfer learning on MS COCO \cite{lin2014microsoft} were performed for YOLOv7. Comparing both results, there are fewer false positives (FPs) in the ROIs for transfer learning such as the omission of the bin and scooter. Higher IoUs are evident for transfer learning out of the True Positives (TPs) identified suggesting a better performance. Most IoU results average between 0.49-0.83 which is a better performance for smaller objects than for typical behavior where smaller objects tend to give lower results in images \cite{zheng2021enhancing}.
%Agnew et al \cite{agnew2023detecting} trained YOLOF on overfilled and non-overfilled bins where inference results were relatively high between the 0.8-0.9 IoU range and the objects are much larger than for Woodscape.
However, the current statistics of YOLO are not meaningful, especially concerning the optical artifacts discussed in this article. Going forward we need an accessible platform on which specific metrics are broken down for each optical artifact, especially in fisheye images.} There are multiple instances of objects being mislabeled by YOLOv7. 
For example, two bicycles placed next to each other or a green bin with four wheels are labeled as `vehicles' in the right background. 
A rack of bicycles is also detected as a vehicle in the left background and the scooter in the right background is incorrectly labeled as `person'. These False Positive errors are likely (in the terminology of Hoiem \cite{hoiem2012diagnosing}) due to similarity or background confusion. Woodscape does not have a `scooter' or a `bin' label as part of the ground truth labeling. 
\ins{Object detection errors of fisheye images are inherently affected by the optical artifacts where the objects in the scenes are deformed as shown in recent literature \cite{chang2019deep, deng2019restricted, tian2021ava, Müller2021}.} Not only must the model deal with optical artifacts discussed previously, but fisheye also introduces rotations of the objects dependent on the scene positions. Examine the bicycles in Figure~\ref{fig:yolo-mvl}. All are approximately vertical in the scene (i.e., they are standing on the ground), but in the fisheye image, they appear in a range of different orientations. Any model must learn rotational invariance to perform well on fisheye surround-view cameras.

\begin{figure}[htb]
     \centering
     \includegraphics[width=0.8\columnwidth]{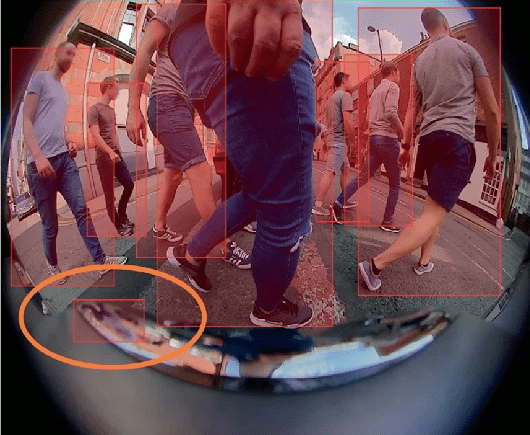}
    \caption{{\bf False positive detected on the vehicle bodywork} due to reflections visible in the surround-view camera \cite{hogan2023automatic}.}
    \label{fig:bodyworkfp}
\end{figure}

In addition to optical and geometric effects, a practical limitation of using fisheye cameras on a vehicle is that much of the image will be composed of the bodywork of the ego-vehicle itself, potentially causing image reflections to be mistaken as objects in the scene (e.g., see Figure~\ref{fig:bodyworkfp}). Recently,  a simple U-Net \cite{Ronneberger2015} architecture with a Res-Net50 \cite{he2016deep} encoder was proposed to extract areas of the vehicle ego-body in the surround-view images of Woodscape to prevent unnecessary computation and detection of false positives \cite{hogan2023automatic}. 

Braun et al. \cite{Braun2022} discuss the difficulty in accounting for all possible scenarios with automotive cameras. Feasibility is one of the main obstacles to building robust computer vision models for the industry where data needs to be collected in real-time to account for the consistently changing environmental conditions and traffic scenarios that have not been tested during development. There are large datasets including BDD100k \cite{bdd100k} and KITTI \cite{geiger2013vision} covering a sample of potential road traffic events on roads but the data in these represent just a sample of possible events in the automotive context. There may be completely different conditions or exceptions such as the time of year when festivals disrupt the expected flow of traffic or unplanned roadworks which adds to the unpredictability of traffic. In short, it is not possible to test all road traffic scenarios using only real-world datasets that have a limited size. For this reason, in Section~\ref{sec:sim}, we will discuss the current state of surround-view simulation systems. First, however, we will briefly discuss some newly developed metrics for computer vision that take into account the spatial variance of image and scene.

\subsection{New Spatially Variant Performance Metrics}
\label{sec:spm}
As mentioned previously, PSFs are used to model the degree of blurring in images created by an optical system. The PSFs can represent the way the optical resolution of the camera system changes from the center to the periphery of an image. However, measuring PSFs is not straightforward. As mentioned, the PSF is essentially the spatial impulse response of the camera. To measure the PSF, you need a scene with specific hardware (such as lasers or specific scene targets) \cite{7299130}. A simpler and more practical measure of spatial resolution is to use the modulation transfer function (MTF) \cite{Schroeder1981mtf}, which is a measure of the spatial frequency response of the camera. Indeed, there are mechanisms to extract the MTF of natural scenes without the need for any specific hardware or target \cite{Zwanenberg2021nssfr}.

The correlation between MTF and computer vision performance has been investigated in \cite{Muller2022}, where it is concluded that MTF is not a suitable predictor for the performance of computer vision. The test results show very little difference in optical performance characteristics using a Cooke Triplet despite the use of completely different sampling methods (i.e., superposition (SP) and isoplanar (ISO) with blocks of area 320 pixels$^2$) \cite{Muller2022}.

A potential strategy identified in \cite{Muller2022} is the use of spatial metrics \cite{brummel2022spatial, Müller2021}. Spatial metrics are a useful approach to quantifying computer vision algorithms, as they report performance measures in different regions within the image. This is especially interesting given the spatially variant optical aberrations exhibited in fisheye cameras. The two key metrics proposed are the Spatial Recall Index ($SRI$) and the Spatial Precision Index ($SPI$) as described by (\ref{eq:SRI}) and (\ref{eq:SPI}) respectively. These are spatial variants that generalize the statistical metrics of Recall and Precision, as the names indicate. In \cite{Muller2022}, both metrics were adapted to quantify the spatial performance of a Cascade Mask RCNN model \cite{cai2018cascade} for instance segmentation in automotive scenes. In each case, the statistical metrics were adapted by assigning the equivalent statistical value (i.e., both Precision and Recall) to each pixel occupied by the object in the bounding box. Using the metrics from (\ref{eq:SRI}) and (\ref{eq:SPI}) a correlation was observed between the spatially varying optical performance and the spatial performance of the Cascade Mask RCNN model. The $SRI$ is given as:
\begin{equation}
\label{eq:SRI}
\begin{aligned}
    SRI = \left[\sum_{n=1}^{N}\left\{ \begin{array}{cl}
1 & (x,y)\in[{ TP}_n\cap P_n] \\
0 & else
\end{array} \right. \right] \\
\oslash\left[\sum_{k=1}^{K}\left\{ \begin{array}{cl}
1 & (x,y)\in[P_k] \\
0 & else
\end{array}\right. \right]
\end{aligned}
\end{equation}
Where, $(x,y)\in[{ TP}_n\cap P_n]$ are all the pixels belonging to the $n$-th true positive mask and the $k$-th ground-truth mask denoted by $P_{k}$. The element-wise division denoted by $\oslash$ of the sum of all TPs and the sum of all ground truths results in the $SRI$ metric.

Similarly, the $SPI$ metric is found by element-wise division of $N$ true positive instances and the sum of $N$ true positive and $J$ false positive instances. 
\begin{equation}
\label{eq:SPI}
\begin{aligned}
    SPI = \left[ \sum_{n=1}^{N}\left\{ \begin{array}{cl}
1 & (x,y)\in[{TP}_n\cap P_n] \\
0 & else
\end{array} \right. \right] \\
\oslash
\left[ 
\begin{aligned}
\sum_{n=1}^{N}\left\{ \begin{array}{cl}
1 & (x,y)\in[{TP}_n] \\
0 & else
\end{array} \right. \\
+ \sum_{j=1}^{J}\left\{ \begin{array}{cl}
1 & (x,y)\in[{FP}_j] \\
0 & else
\end{array} \right.
\end{aligned}
\right]
\end{aligned}
\end{equation}
Where, $(x,y)\in[{ FP}_j]$ denotes the $j$-th false positive instance and $(x,y)\in[{ TP}_n]$ denotes the $n$-th true positive instance.
Notice the same numerators are used in both (\ref{eq:SRI}) and (\ref{eq:SPI}). For more information please refer to the original work \cite{brummel2022spatial, Müller2021}.

These spatial metrics were used on degraded versions of the BDD100k datasets where a defocus of $\Delta{z}=\pm{1.25}$ was applied. Differences in spatial performance were examined by comparing the results from the degraded datasets to the baseline BDD100k dataset. This is known as the spatial performance drop ($SRI_{drop}$ and $SPI_{drop}$):
\begin{equation}
    \label{eq:SRId}
    SRI_{drop} = SRI_{base} - SRI_{\Delta{z}}
\end{equation}
\begin{equation}
    \label{eq:SPId}
    SPI_{drop} = SPI_{base} - SPI_{\Delta{z}}
\end{equation}
Where $SRI_{base}$ and $SPI_{base}$ denote the spatial results of the baseline dataset and $SRI_{\Delta{z}}$ and $SPI_{\Delta{z}}$ denote the spatial results of the degraded datasets.

Four observations can be made from the experiments in \cite{brummel2022spatial, Müller2021}:
\begin{enumerate}
\item{It was found that AP dropped from 87.21\% (i.e. baseline) to 84.78\% for a defocus of $\Delta{z} = +1.25$ and 82.72\% for $\Delta{z} = -1.25$, respectively.}
\item{Spatial performance drop increases towards the periphery or edges of the images for a negative defocus (i.e. $\Delta{z} = -1.25$). Conversely, performance drop increases towards the center of the images for a defocus of $\Delta{z} = +1.25$ suggesting that statistical metrics vary across the FOV.}
\item{Ground truth instances were missed due to the worsening image quality between $\Delta{z}=\pm{1.25}$.}
\item{Finally, the number of FPs was not influenced negatively by the degradations.}
\end{enumerate}
Both of these works demonstrate the additional insight that can be obtained from these metrics and should become part of the analysis when considering optical artifacts in fisheye images.

\section{Surround-View Simulations}
\label{sec:sim}

Now we will turn our attention to surround-view simulation. \ins{As a way of resolving the lack of understanding of combining fisheye cameras with computer vision (especially due to the lack of public datasets available); automotive simulation can be utilized. However, as will be discussed in this section, automotive simulation lacks the optical properties of the camera and there is no standardized process of perception algorithms for automotive simulation.} As we have already discussed, datasets will, no matter how large they are, suffer from a limited set of scenarios in which they are captured. Additionally, the collection of real-life data for training AI applications in the automotive industry (including autonomous driving) is a time-consuming and expensive process \cite{kumar2023surround, carlson2018modeling}. The BDD100k dataset \cite{bdd100k} has over 100k driving videos collected from more than 50k rides capturing New York and the San Francisco Bay area (among others) and is one of the largest automotive datasets released to date. Woodscape \cite{yogamani2019woodscape} has 10k publicly released images from three distinct locations (USA, Europe, and China) and is the first dataset of its kind to address surround-view cameras for automated driving. These datasets, while giving a reasonable amount of data for training computer vision algorithms for research, have their limitations where they only show a fraction of what can happen in the real world. Thus, it is fruitful to provide an overview (and some discussions) of current trends in automotive scene simulations, including model-driven and data-driven simulators, which are proposed to be used for addressing corner cases in autonomous driving. In the context of simulation, we also revisit the optical artifacts from Section~\ref{sec:optical-effects} and discuss how they should be combined with the fisheye distortion models for photorealism.
\begin{figure*}[t]
    \centering
    %befor width = 4.5in 
    \includegraphics[width=0.7\textwidth]{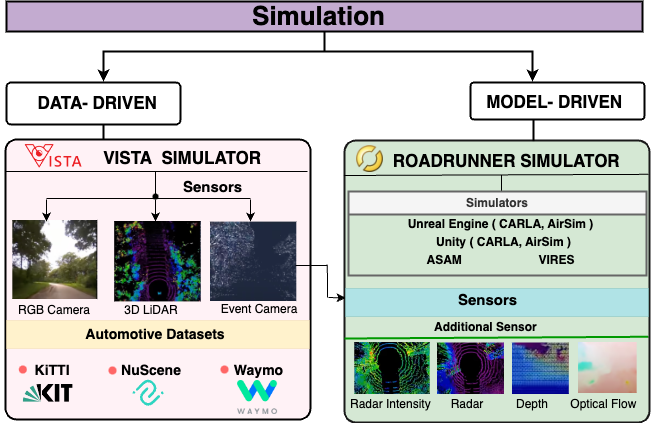}
    \caption{{\bf Simulation Hierarchy}. MATLAB RoadRunner custom maps can be exported to any model-driven simulator. Unlike model-driven simulators, VISTA uses real-life datasets (e.g., KITTI, NuScenes, and Waymo) to generate simulated images or additional viewpoints of the same environment. Also, the sensors available in the VISTA simulator (i.e., RGB Camera, 3D LiDAR, and Event Camera) can also be found in the model-driven simulators.}
    \label{fig:sim-hier}
\end{figure*}
\begin{figure*}[t]
     \centering
     \includegraphics[width=7in, keepaspectratio]{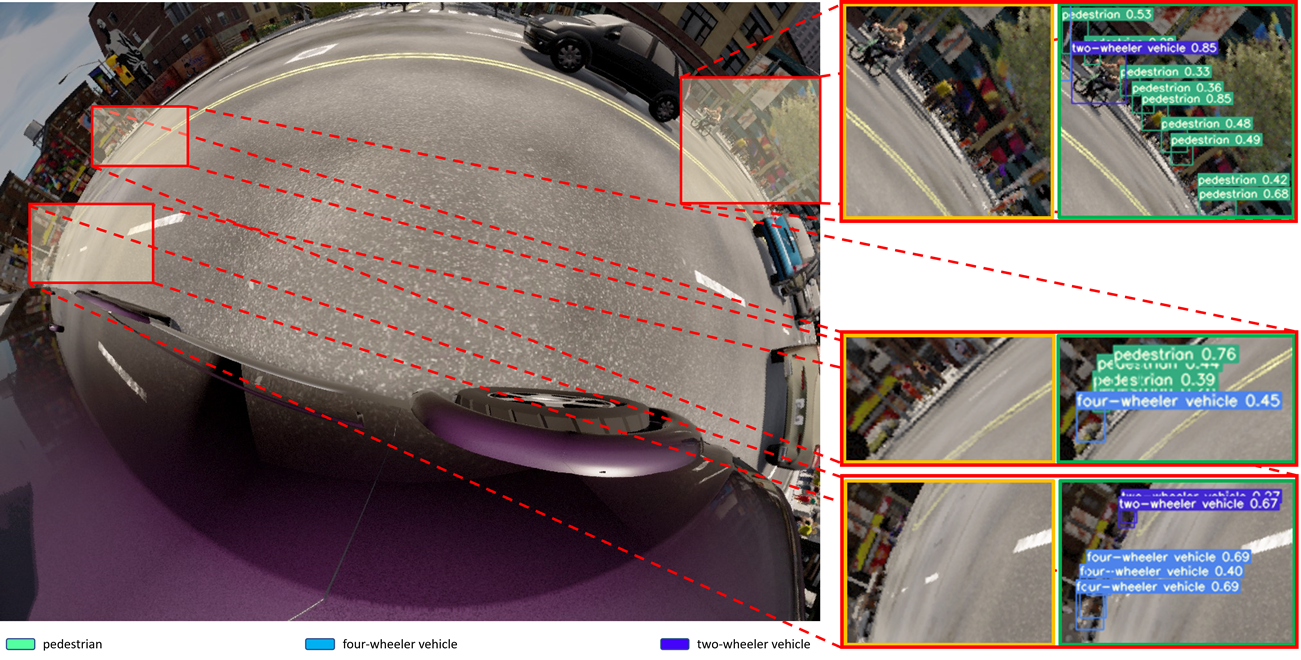}
    \caption{\ins{{\bf YOLOv7 trained from scratch on SynWoodscape and inference on an MVL sample (green box)}. Zoomed areas of the image show regions of interest (ROIs) for object detection (red box). See Figure~\ref{fig:yolo-mvl} for comparison. Note the absence of mechanical vignetting (dark corners), artificial lighting (uniform colors), and the presence of the outline of the cube map fisheye projection model used for distortion (a dark outline of a cube in the center of the image.}}
    \label{fig:synyolo-mvl}
\end{figure*}
\subsection{The Argument for Simulations}

\ins{As stated previously simulation lacks the optical properties of the camera but can potentially be considered as a direction of research for integrating the camera design with the simulation tools available.}
It is quite intuitive why simulations are needed for the development of automated driving solutions. However, it is still worth taking the time to make this explicit (perhaps for the uninitiated reader). Every time a new traffic scenario is found that is not covered by the public datasets discussed (or other datasets), the new information should be added and re-split into the given ratios of training, validation, and testing.  For example, if new data was added to the Woodscape dataset in the experiment from Section \ref{sec:cv-fisheye} Figure~\ref{fig:yolo-mvl}, the data would need to be re-split into the (80\%-10\%-10\%) ratio and YOLOv7 would need to be retrained. This would be very inefficient in the long term, especially if a diverse set of camera models with different lens calibrations would need to be incorporated into the datasets.
Therefore, by adding simulation data to real-world data, \ins{the performance of computer vision can improve on real-world data. This can also be possible where there is a lack of ground-truth data in real-world images for training computer vision. For example, as of the writing of this survey, SynWoodscape \cite{Sekkat2022} is the only publicly available synthetic fisheye dataset with pixel-wise optical flow ground truth. For the original Woodscape, it was not possible to obtain optical flow ground truth in the real scenes, and synthetic data was identified as the only feasible solution \cite{yogamani2019woodscape}. As a logical next step, Shen's experiments \cite{shen2023optical}, demonstrate using the ground truth from SynWoodscape on the Recurrent All-Pairs Field Transforms (RAFT)\cite{teed2020raft} optical flow algorithm to improve performance and correct erroneous optical flow in the original Woodscape. Table \ref{tab:results} shows that synthetic data can be considered to simulate optical artifacts due to the realism of automotive simulators. However, as clearly stated in this section, there are quantitative performance differences between degraded simulation data and real-world data. Therefore, a more comprehensive analysis of computer vision algorithms is required and any simulation results shown here by no means represents a definitive analysis. Simulation must develop in the direction of combining the optical properties of the camera with simulation for there to be any comprehensive analysis.}

\ins{
However, the lack of real-world optical artifacts inside the simulation presents a disadvantage when it comes to optimizing algorithms for the real world. In the context of model-driven simulators, the only approach to solving this problem is by implementing the lens design inside the simulator which would lead to the inherent addition of optical artifacts and a more realistic synthetic image than before. One such strategy would be combining the well-known OpticStudio\textsuperscript{\ref{ops}} with CARLA \cite{Dosovitskiy2017} or Unity where the ability to consistently alter lens models gives flexibility to the user as a way of varying optical artifacts inside the simulator.}

As of this moment, data augmentation with pixel unprojection (to a ray), rotation, and re-projection (of the ray to pixel) of the fisheye data can be used as a means of creating augmented fisheye images for training computer vision algorithms \cite{konrad2021fisheyesuperpoint, cho2023surround}. However, creating augmented fisheye data affects the distribution of optical artifacts in the images in a possibly unrealistic manner. Additionally, by using the original images for augmentation, no new lens configuration can be learned from the augmented data, as the augmentation just models a warped version of the lens configuration of the original system.
%In most cases, there are more dark regions in which the original fisheye images are disproportionately scaled. This does not represent new data that would be suitable for examining optical performance. 
Therefore, augmented fisheye data is not necessarily a solution to examining optical performance behavior on computer vision. If the original data were to be synthesized (or augmented) for training, both the original and altered datasets would only ever have one specific lens configuration. This could potentially lead to overfitting computer vision on one specific lens configuration, which is not useful for scaling to different cameras.

The naive approach to solving the data problem would be to iterate over an entire set of camera lenses. This is impossible for three reasons:
\begin{enumerate}
\item{Firstly, using only one camera is not enough to test all possible combinations. Every time a new cycle of experiments begins the camera would need to be disassembled and reassembled with new lens calibrations, which would be a time-consuming process with potentially irreproducible results.}
\item{Secondly, it is infeasible to test all possible lens calibration combinations for computer vision where data is collected with a group of cameras with different optical configurations. This is mainly because it is impossible to determine the number of cameras needed for this experiment without making mistakes in assumptions, for example, about the optical artifacts discussed in Section~\ref{sec:optical-effects}.}
\item{Finally, it would likely only be possible to estimate this number via trial-and-error experiments. This could potentially lead to an unpredictable number of iterations in experiments to demonstrate a smooth transition of results iterating over the different lens designs.}
\end{enumerate}

A solution to this problem is to use simulation by varying optical parameters of lenses and examine the effect on computer vision performance. As of today's research, automotive simulation can be broken down into the hierarchy presented in Figure~\ref{fig:sim-hier}, where simulation can be both model-driven and data-driven. 

\subsection{Model-Driven Simulators}
\label{sec:model-d}
%\begin{figure}[h]
%     \centering
%     \includegraphics[width=0.8\columnwidth]{pictures/F2BEV_Volvo.png}
%     \caption{{\bf A sample from FB-SSEM FV Volvo Unity Dataset} \cite{samani2023f2bev}. In this case, a native model for fisheye is available. Hence, the artifact of the projection cube is not visible. However, we can see that other optical artifacts that should be present to model an actual camera are not.}
%     \label{fig:F2BEV_Volvo}
%\end{figure}
Model-driven simulation is where simulation is created using computer programs to design a scene and landscape using blueprint assets and mesh vectors. As described by Mutsche et al. \cite{mutsch2023model}, model-driven simulation is more comparable to modern video games. As shown in Figure~\ref{fig:sim-hier}, there is a wide range of model-driven simulators available. Each simulator is designed with sensors or multi-modal outputs (i.e., camera, radar, Light Detection and Ranging (LiDAR), etc.) applicable to autonomous driving. Model-driven simulation is challenging in automated driving mainly because it represents an artificial and near-perfect version of real life, which does not actually exist. There is a visible difference in texture used on assets, buildings, and the environment.
For example, the asphalt on the road as depicted in Figure~\ref{fig:synyolo-mvl} is unnaturally brighter than the road surface in Figure~\ref{fig:yolo-mvl}, clearly indicating that, while objects may resemble real life, current automotive sensor simulators struggle with texture. Despite these limitations, simulation is improving rapidly in terms of visual quality. The recent release of Unreal Engine 5.2 (UE5) has much more control over the look and feel of objects in scenes, such as applying dust on the clear coating of vehicles. 
These visual improvements suggest that model-driven simulation is steadily advancing towards photorealism, at least superficially.
By utilizing automotive simulators, such as CARLA \cite{Dosovitskiy2017}, scarcity of data issues are addressed in datasets for supervised autonomous driving. Therefore, if there were a way to combine real and synthetic data, it would be possible to measure optical performance across a range of lens calibrations to understand which lens combination would be ideal for which computer vision model. Especially in the space of surround-view cameras, a simulator such as CARLA with Unreal Engine \cite{Dosovitskiy2017} would be useful to test out these different scenarios in combination with fisheye distortion. However, CARLA cannot natively model fisheye cameras. 

\begin{figure}[t]
     \centering
     \begin{subfigure}{0.49\columnwidth}
         \centering
         \includegraphics[width=0.9\columnwidth]{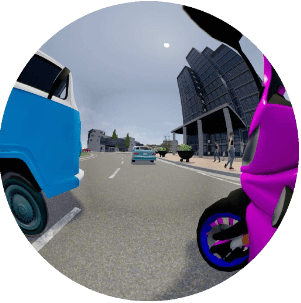}
         %\caption{OmniScape Left}
         %\label{fig:omniscape-left}
     \end{subfigure}
     \begin{subfigure}{0.49\columnwidth}
         \centering
         \includegraphics[width=0.9\columnwidth]{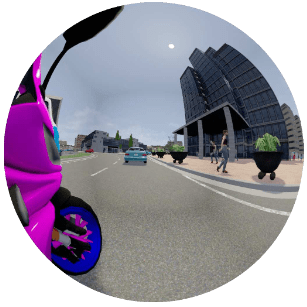}
         %\caption{OmniScape Right}
         %\label{fig:omniscape-right}
     \end{subfigure}
        \caption{{\bf Synthetic fisheye images from the OmniScape dataset left and right views} \cite{Sekkat2020omniscape}. Note the visibility of the projection cube as shadows in the image. }
        \label{fig:synfisheye}
\end{figure}

\begin{figure}[h]
    \centering
    \includegraphics[width=0.8\columnwidth]{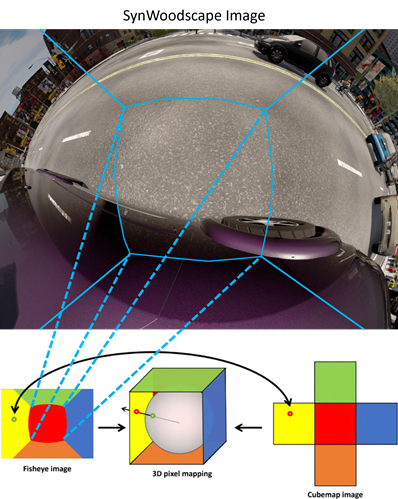}
    \caption{\ins{{\bf Mapping of the cube map image’s pixels to the fisheye image} \cite{Sekkat2022}. Five or six perspective cameras are created in the simulator. Each side of the cube map has a 90$^\circ$ FOV. These are then post-processed into a fisheye image.}}
    \label{fig:cubemap}
\end{figure}

To overcome this, to the best of the authors' knowledge, all fisheye models created in CARLA use an intermediate cube map for the creation of the fisheye image \cite{Sekkat2020omniscape, Sekkat2022, berenguel2020omniscv, ang2020big}, shown in Figure~\ref{fig:cubemap}. The cube map can then be mapped to a fisheye image using any fisheye model. For example, in \cite{Sekkat2020omniscape}, a 10,000 image dataset of fisheye from CARLA was converted using a six cubic map model, as shown in Figure~\ref{fig:synfisheye}. The cube map images were converted to fisheye using the model from \cite{scaramuzza2006toolbox}. The SynWoodscape dataset took a similar approach, except using the fourth-order polynomial model from \cite{yogamani2019woodscape}. However, a major drawback of this two-step fisheye image creation is that, due to the initial cubic mapping, there are faint dark outlines of the cube map projection model itself, such as can be seen in OmniScape (Figure~\ref{fig:synfisheye}, and in SynWoodscape (Figure~\ref{fig:cubemap}).

\ins{Despite the lack of realism in synthetic data, YOLOv7 does surprisingly well at identifying objects in SynWoodscape (split into an 80\%:10\%:10\% ratio) which has the same degree of distortion as Woodscape (see Figure \ref{fig:synyolo-mvl}). Most notably, even with a relatively small dataset of 2.5k images, it is capable of identifying very small pedestrians with the IoU threshold set to 0.25. %With the addition of optical artifacts into the simulation as demonstrated by Carlson et al. \cite{carlson2018modeling} the performance of computer vision algorithms could improve overall on real world data.
}
A recently released synthetic automotive surround-view dataset, called Fisheye camera images and BEV maps from Simulated Sequences of Ego car Motion (FB-SSEM) \cite{samani2023f2bev}, was built using the Unity game engine. This dataset is interesting, as it has both Depth BEV and fisheye segmentation views, which can be used for a wide range of computer vision algorithms such as object detection, segmentation, and trajectory prediction (i.e., predicting the position of the ego-vehicle in a scene).
%An example of an FV perspective is shown in Figure~\ref{fig:F2BEV_Volvo}. 
As the Unity game engine has built-in fisheye support, it is interesting to note that we don't have the same artifacts that occur with the cube map approach in CARLA. However, this adds a restriction that, in the Unity game engine, one can only model a fisheye camera using the projection model implemented in the Unity source code. As previously discussed, there are many models of fisheye projection that one may be interested in \cite{kumar2023surround}.

A diverse set of maps is required for an algorithm to learn more efficiently, as information collected from only one map or scene has a bias for this specific traffic scenario. RoadRunner\footnote{RoadRunner Map Editor: \url{https://uk.mathworks.com/products/roadrunner.html}} offers a solution that is a Map Editor tool designed to create automotive maps for simulators like CARLA or Unity. With this tool, researchers can design their custom maps to suit their applications in autonomous driving. Using maps with different props, actors, and scenery can also widen the possibilities of fisheye dataset creation.

\subsection{Data-Driven Simulators}
\ins{Data-driven simulation is a relatively unexplored area in the simulation field where instead of designing computer models that imitate the realism of automotive scenarios, publicly gathered datasets (such as the KITTI \cite{geiger2013vision}, NuScenes \cite{caesar2020nuscenes}, and Waymo \cite{Sun_2020_CVPR} datasets) can be synthesized for simulation which would have a more photorealistic appearance. A publicly available interface is being constructed to interface with these three datasets\footnote{Interface to public datasets available at: \url{https://vista.csail.mit.edu/interface_to_public_dataset/index.html\#interface-to-public-dataset}}. As of the writing of this paper, the interface for these public datasets has not yet been released on the official website.
The second version of Virtual Image Synthesis and Transformation for Autonomy (VISTA 2.0) \cite{amini2022vista} has the current capabilities to accommodate different sensor systems such as RGB, LiDAR, and Event-based cameras.}

The lack of photorealism in simulators like CARLA raises the question as to whether model-driven simulation is reliable for development and testing in the long term, despite its practicality and feasibility. This is where photorealistic simulation, such as VISTA 2.0 can potentially help \cite{amini2022vista, Amini2020}. VISTA is a data-driven simulator, which can adapt its actors and virtual environment to that seen in real-life datasets. Data-driven simulation is a relatively new area of simulation development which, as yet, is unproven in surround-view imaging for automated driving.

The adaptation of public datasets such as KITTI, will provide an advantage in creating custom synthetic datasets following these benchmark implementations. For example, creating a synthetic fisheye dataset using Woodscape with VISTA would be a convenient way of generating synthetic fisheye data without learning the complexity of model-driven simulators. This would be especially useful to target scenarios not covered by the traditional simulators and improve the overall quality of testing. However, for this paper, it is important to note that, at the time of publication, there seem to be no plans to convert any surround-view dataset with VISTA.

The benefit of VISTA is that it can generate an infinite set of local viewpoints of trajectories from only a sparsely sampled set of real-world viewpoints of a scene. Each newly generated viewpoint is a different view of the same scene. In Amini's experiments \cite{Amini2020}, VISTA outperformed three other self-driving techniques with viewpoint augmentation: (1) Domain Randomization \cite{tobin2017domain}, (2) Sim-to-Real Domain Adaption \cite{bewley2019learning}, and (3) Imitation Learning \cite{bojarski2016end}. Unlike the other self-driving strategies, VISTA required no lane interventions, and the near-crash recovery rate improved by 30\% when compared to Imitation Learning.

The VISTA algorithm works in the following manner \cite{amini2022vista}:
\begin{enumerate}
    \item{In the ego-vehicle's state, a continuous kinematic model is updated consisting of the virtual agent's dynamics and those of the human driver.}
    \item {The closest frame concerning the vehicle’s pose is taken and a scene is recreated based on the ego-vehicle's perspective.}
    \item{The frame is projected into 3D space to reconstruct the scene.}
    \item {From the 3D space a coordinate transformation is performed using the virtual agent's viewpoint.}
    \item{A new 2D simulated observation predicting the next frame of the scene is reprojected from the 3D transformation back as a new viewpoint.}
    \item{The newly generated observation is sent back to the agent which then decides the ego-vehicle's next course of action in the scene. For example, if the road veers to the right, the driving wheel is turned to the right.}
\end{enumerate}

While data-driven simulation looks powerful, even if in the early stages, there are still some obvious artifacts. For example, if one carefully examines the examples provided in \cite{amini2022vista}, under a modeled camera translation, the resulting images have unrealistic perspective distortions introduced.
\subsection{Optical Artifacts in Simulation}
\begin{figure}[t]
    \centering
    \includegraphics[width=0.9\columnwidth]{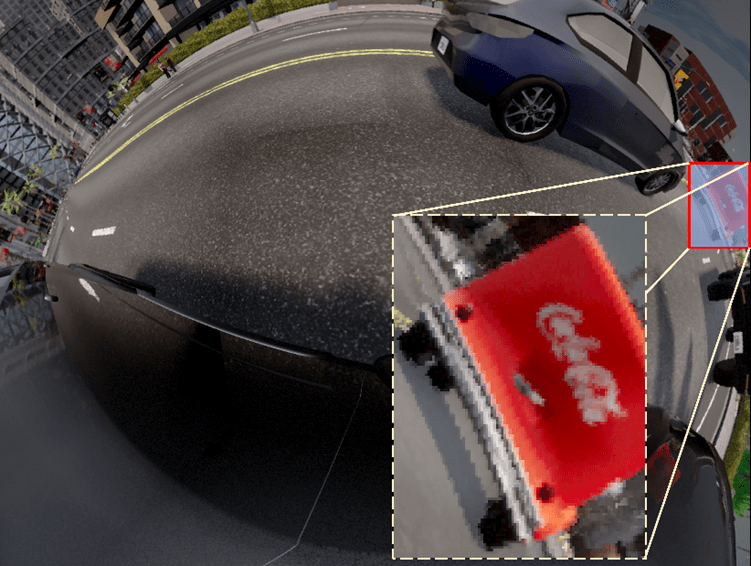}
    \caption{{\bf SynWoodscape left camera image showing no evidence of lateral or longitudinal chromatic aberration}. See Figure~\ref{fig:z_W} for comparison. Note also that the projection cube is visible as a shadow structure in the image.}
    \label{fig:syn_no_CA}
\end{figure}

While simulators produce data that can enhance performance for computer vision, questions remain as to how effectively lens models can translate to a simulation where there is no lens present in the virtual world. Of course, data-driven simulators pose an interesting question as to whether lens configuration is transferable from real datasets, such as KITTI. However, as clearly indicated in SynWoodscape Figure~\ref{fig:syn_no_CA}, there is no evidence of optical artifacts such as chromatic aberration, astigmatism (or, in fact, any significant PSF), or vignetting. These simulators are not designed to incorporate real-world optics and can often bring their own artifacts (like the visibility of the projection cube).

In Carlson et al. \cite{carlson2018modeling} optical artifacts are accounted for in computer vision experiments by applying them using filters (such as chromatic aberration) to the Virtual KITTI \cite{gaidon2016virtual} and Grand Theft Auto (GTA) \cite{richter2016gta} datasets.
\begin{figure}[t]
    \centering
    \includegraphics[width=0.9\columnwidth]{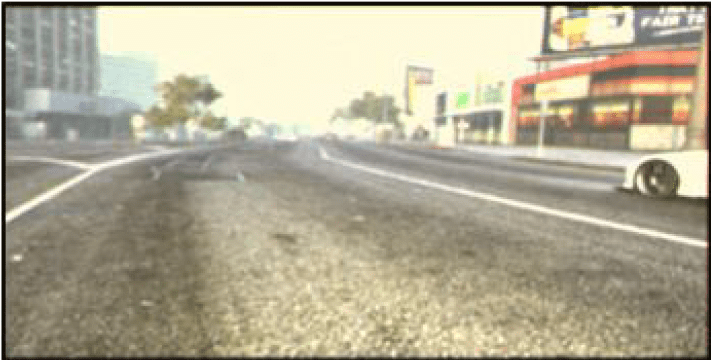}
    \caption{{\bf GTA dataset with optical artifacts applied} \cite{carlson2018modeling}.}
    \label{fig:gta-blur}
\end{figure}
A sample illustration of the resulting image from GTA can be seen in Figure~ \ref{fig:gta-blur}. It was shown in \cite{carlson2018modeling}, from the results with baseline (unaugmented) and augmented (i.e., synthetic images with chromatic aberration, blur, noise, etc.), that by adding the optical artifacts uniformly across the synthetic images in training, Faster Region Convolution Neural Network (RCNN) models performed better with slightly improved mean Average Precision (mAP) compared to the unaugmented synthetic datasets (see Table \ref{tab:results}). Questions remain unanswered as to how to control the application of these optical artifacts in the spatial domain, especially for fisheye lenses where strong optical aberrations are not uniform across the image.

In summary, while graphical simulation may seem like a feasible solution to cover unforeseen scenarios on roads, it is not the overall solution as currently, simulation alone cannot completely replace the realism of actual camera footage \cite{Braun2022}.

\section{Conclusion}
\label{sec:conclusion}
%- reiteration of domain and problem
%- repeat that we did survey/review on optical effects
%- reiterate limitations found in simulation
%- what are future steps in field to address limitations

%- reiteration of domain and problem
%- repeat that we did survey/review on optical effects
Automotive surround view cameras have become a niche research area for computer vision. This survey has identified the gaps and limitations of current strategies to train computer vision systems for wide FOV cameras, both in real life and in simulation. In particular, this survey has discussed fisheye optical systems, the optical effects that inherently affect these systems, and the concept of adding optical effects in simulation.

Concerning computer vision performance in surround-view systems, there is little work in the space that examines the impact of optical artifacts. Some computer vision tasks, such as those that require depth extraction, appear to be improved when certain artifacts are present. One might think, therefore, that for these tasks designing a camera system with a significantly greater presence of these artifacts may be beneficial. However, this is not indicated by the results in this area, in which only residual artifacts are examined. Likely, the imperfections due to these optical artifacts would quickly damage the performance of these tasks, were they too great. Regardless, other tasks are almost certainly negatively affected by optical artifacts. What is clear is that significantly more work in this space is required for the community to understand completely the impact of optical artifacts on computer vision performance, particularly for surround-view fisheye cameras, where design challenges mean these artifacts can be significant.

%- reiterate limitations found in simulation
% identify other limitations
From this survey, two limitations in surround-view simulators can be identified.
Firstly, \textbf{popular simulators lack the simulation of lenses from the real world}. Geometric distortion models have been applied in simulation to introduce fisheye distortion. However, real fisheye images are not just affected by distortion but by a wide range of optical effects, as has been discussed. Also, it is only possible to apply optical effects to simulated images external to simulators, which is a problem as there is little to no chance of reproducibility for projects even if it can be achieved indirectly outside of simulators.
Secondly, as identified in the discussions from Section \ref{sec:sim}, there is a \textbf{scarcity of both real and simulated fisheye datasets for automated driving}. This hinders the development of surround-view perception systems, where current simulation data may prove unsuitable for training, partially due to the lack of optical effects.

%\section*{Acknowledgment}
%The authors would like to sincerely thank Lero Science Foundation Ireland and Valeo Vision Systems for making this research possible.

\bibliographystyle{IEEEtran}
\bibliography{bibliography-ieee}

\begin{IEEEbiography}[{\includegraphics[width=1in,height=1.25in,clip,keepaspectratio]{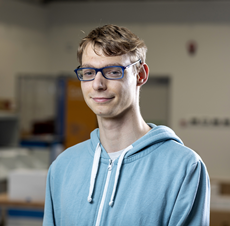}}]{\textbf{Daniel Jakab}}received the M.E. degree in Electronic and Computer Engineering from the University of Limerick, Ireland in 2022. He previously worked for 9 months in an internship with Black Box R\&D between (2019-2020). He is currently pursuing a full-time PhD degree with the Department of Electronic and Computer Engineering, University of Limerick, Ireland. His Ph.D. is under the sponsorship of Lero Science Foundation Ireland-funded project, 'Treating fisheye cameras natively using CNNs to improve production tolerances of cameras'.
\end{IEEEbiography}
\begin{IEEEbiography}[{\includegraphics[width=1in,height=1.25in, clip,keepaspectratio]{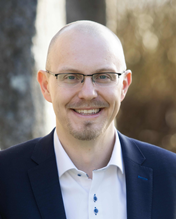}}]{\textbf{Brian Deegan}} received a Bachelor's Degree in Computer Engineering from the University of Limerick in 2004, an MSc In Biomedical Engineering from the University of Limerick in 2005 and a Ph.D. in Biomedical Engineering from the University of Galway in 2011. The focus of his research was the relationship between blood pressure and cerebral blood flow in humans.
From 2011 to 2022 Brian worked in Valeo Vision Systems as a Vision Research Engineer focusing on Image Quality. Brian's research focus is on high dynamic range imaging, LED flicker, Topview harmonization algorithms, and the relationship between image quality and machine vision.
In 2022 Brian joined the Department of Electrical \& Electronic Engineering at the University of Galway as a Lecturer and Researcher.
\end{IEEEbiography}
\begin{IEEEbiography}[{\includegraphics[width=1in,height=1.25in,clip,keepaspectratio]{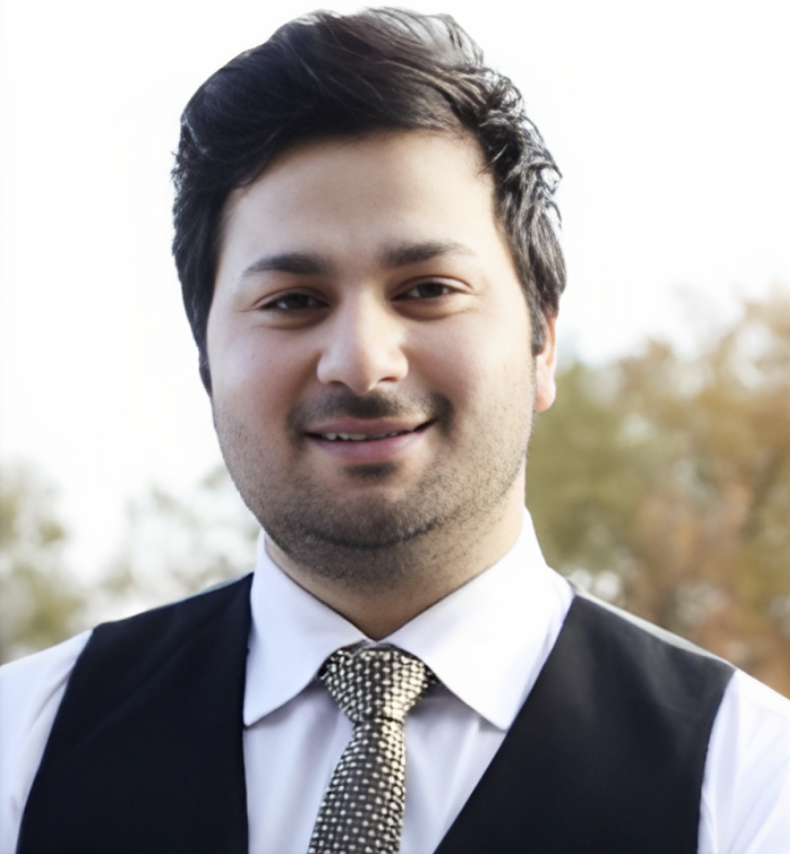}}]{\textbf{Sushil Sharma}}  received his B.Tech. degree in Electronics \& Instrumentation from ITM University, India, in 2014. He completed his M.Sc. degree in Automatic Control and Robotics, specializing in Smart Aerospace and Autonomous Systems, from Université Paris-Saclay, France, and Poznan University of Technology, Poland 2017.
Currently, he is pursuing a full-time PhD degree with the Department of Electronic and Computer Engineering at the University of Limerick, Ireland. His research interests in "Automotive Parking Systems" involve a broad range of areas such as MultiCamera Perception, trajectory prediction, and Machine-Based Algorithms.
\end{IEEEbiography}
\begin{IEEEbiography}[{\includegraphics[width=1in,height=1.25in,clip,keepaspectratio]{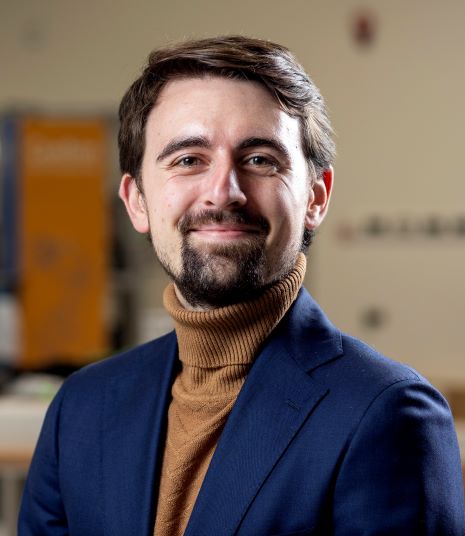}}]{\textbf{Eoin M. Grua}} was born in 1993 in Cork, Ireland. He received the B.S. degree in liberal arts and sciences from Amsterdam University College, Amsterdam, The Netherlands, in 2015 and an M.S. degree in Computer Science from Swansea University, Swansea, Wales, in 2016. In 2021 he received his Ph.D. degree in computer science from the Vrije Universiteit Amsterdam, Amsterdam, The Netherlands. In 2021 he was a research assistant at the University of Limerick, Limerick, Ireland. He is currently a PostDoc in the Department of Electronic and Computer Engineering at the University of Limerick, Limerick, Ireland. His research interests include artificial intelligence, software engineering and architecture, and sustainability. 
\end{IEEEbiography}

\begin{IEEEbiography}[{\includegraphics[width=1in,height=1.25in, clip,keepaspectratio]{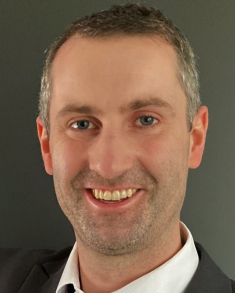}}]{\textbf{Jonathan Horgan}} is a Computer Vision and Deep Learning Architecture Manager and a Senior Expert with Valeo Vision Systems. He has worked in the field of computer vision for over 16 years, with a focus over the last ten years on automotive computer vision for advanced driver assistance systems (ADAS), automated parking, and automated driving. He is currently working on next-generation advanced computer vision and deep learning, with the ultimate goal of achieving fully autonomous driving and parking. He has 25 publications in peer-reviewed journals and conferences and over 100 patents published in the field of automotive computer vision.
\end{IEEEbiography} 
%\vskip -1.5\baselineskip plus -1fil
\begin{IEEEbiography}[{\includegraphics[width=1in,height=1.25in, clip,keepaspectratio]{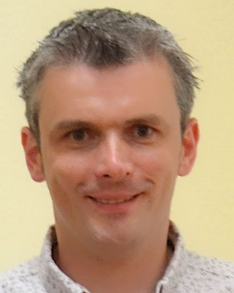}}]{\textbf{Enda Ward}} received the B.E. degree in electronic engineering from the University of Galway, in 1999, and the M.Eng.Sc. degree in electronic engineering, with a focus on biomedical electronics, in 2002. He is responsible for defining the camera product roadmap for surround and automated driving applications with Valeo. He has worked with key technology experts across the supply chain and within OEMs to define optimal system architectures. He lectured for a number of years in the areas of electronics and computing systems with Atlantic Technological University, Ireland. Later, he moved into industry, working in the biomedical space, and has spent the last 16 years in automotive ADAS design. He holds several patents in the area of automotive vision.
\end{IEEEbiography} 
%\vskip -1.5\baselineskip plus -1fil

\begin{IEEEbiography}[{\includegraphics[width=1in,height=1.25in,clip,keepaspectratio]{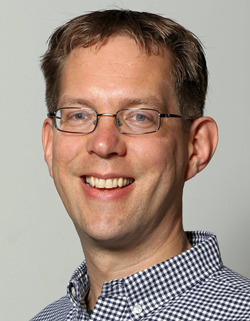}}]{\textbf{Pepijn van de Ven}} received an MSc in Electronic Engineering from the Eindhoven University of Technology (Netherlands) in 2000 and a PhD in Artificial Intelligence for autonomous underwater vehicles from the University of Limerick in 2005. In 2018 he joined UL's teaching staff as a Senior Lecturer in Artificial Intelligence. Pepijn's research interests are in the area of artificial intelligence and machine learning with a particular interest in medical applications. 
\end{IEEEbiography}
\begin{IEEEbiography}[{\includegraphics[width=1in,height=1.25in,clip,keepaspectratio]{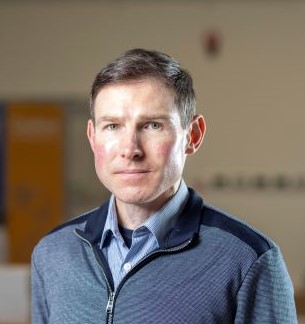}}]{\textbf{Anthony Scanlan}} received the B.Sc. degree in experimental physics from the National University of Ireland Galway, Galway, Ireland, in 1998 and the M.Eng. and Ph.D. degrees in electronic engineering from the University of Limerick, Limerick, Ireland, in 2001 and 2005, respectively. He is currently a Senior Research Fellow at the Dept. of Electronic \& Computer Engineering, University of Limerick, Ireland, and has been the principal investigator on several research projects in the areas of signal processing and data converter design. His current research interests are in the areas of artificial intelligence, computer vision, and industrial and environmental applications. 
\end{IEEEbiography}
\begin{IEEEbiography}[{\includegraphics[width=1in,height=1.25in,clip,keepaspectratio]{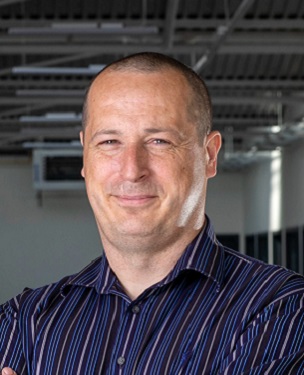}}]{\textbf{Ciarán Eising (Senior Member, IEEE)}} received the BE in Electronic and Computer Engineering and a PhD from the NUI Galway, in 2003 and 2010, respectively. From 2009 to 2020, he was a Computer Vision Architect Senior Expert with Valeo. In 2020, he joined the University of Limerick as an Associate Professor of AI.
\end{IEEEbiography}

\end{document}